\documentclass[11pt]{article}

\usepackage[preprint]{acl-style/acl}

\usepackage{times}
\usepackage{latexsym}

\usepackage[T1]{fontenc}

\usepackage[utf8]{inputenc}

\usepackage{microtype}

\usepackage{inconsolata}

\usepackage{graphicx}

\usepackage{hyperref}
\usepackage{url}
\usepackage{longtable}
\usepackage{cuted}
\usepackage{amsmath}
\usepackage{graphicx}
\usepackage{subcaption}
\usepackage{booktabs}
\usepackage{stfloats}
\usepackage{csquotes}
\usepackage{placeins}

\title{Latent Cluster Analysis for Vision-Language-Action Models}

\author{
 \textbf{Theodor Wulff\textsuperscript{1,*}},
 \textbf{Sergio Lanza\textsuperscript{2,*}},
 \textbf{Tamara Bila\textsuperscript{3,*}},
\\
 \textbf{Angelo Cangelosi\textsuperscript{1}},
 \textbf{Stefan Wermter\textsuperscript{2}},
 \textbf{Igor Farkas\textsuperscript{3}}
\\
 \textsuperscript{1}The University of Manchester,
 \textsuperscript{2}Universität Hamburg,
 \textsuperscript{3}Comenius University Bratislava
\\
 \small{
   \textbf{Correspondence:} \href{mailto:theodor.wulff@manchester.ac.uk}{theodor.wulff@manchester.ac.uk}, \href{mailto:sergio.lanza@uni-hamburg.de}{sergio.lanza@uni-hamburg.de}, \href{mailto:tamara.bila@fmph.uniba.sk}{tamara.bila@fmph.uniba.sk}
 }\\
 \small{
    *equal contribution
 }
}

\begin{document}
\maketitle

\begin{abstract}

Vision-Language-Action (VLA) Models are increasingly used in robotics for their ability to ground language and perception into action, yet the internal representations driving their behaviour remain poorly understood. 
We propose LAVLA, a framework for latent cluster analysis of VLA models, and conduct a layer-wise study of the state-of-the-art GR00T N1.5 model, with particular focus on its action decoder. To better characterise the latent space during action diffusion, we introduce a cross-attention-based embedding-weighting method that amplifies relevant features while suppressing less informative ones. Quantitative evaluation shows that weighted clustering consistently outperforms the baseline. To improve interpretability, we extract human-interpretable concepts for each cluster, linking latent representations to semantic descriptions. Our analysis shows that latent clusters progressively disentangle spatiotemporal and kinematic features, with representations becoming more refined in the middle layers and stabilising toward the output.
As such, LAVLA advances the interpretability of language-driven robotic systems.
\end{abstract}

\section{Introduction}
Vision-Language-Action Models (VLAs) are gaining popularity in robotic research due to their promising generalisation capabilities.
Conventionally, VLAs are trained on large-scale collections of robot data across several environments, tasks, and embodiments, enabling strong performance in different scenarios and architectures~\cite{kim2025openvla,nvidia2025gr00tn1openfoundation}.
While performance continues to improve as the number of environments and considered tasks increases, only a limited amount of research is concerned with explaining the inner reasoning of these models~\cite{haonMechanisticInterpretabilitySteering2025a}.
Extending the understanding of these models is a key factor in robotics, where the range of possible out-of-distribution scenarios is vast, and unverified behaviour can have dire consequences.
Gaining insight into how models encode multimodal information improves user trust and facilitates the identification of reasoning failures~\cite{ma2024doesvlmclassificationbenefit,zang2025pretrainedvisionlanguagemodelslearn}.
Based on these premises, we introduce a novel framework for \textbf{L}atent Cluster \textbf{A}nalysis for \textbf{V}ision-\textbf{L}anguage-\textbf{A}ction Models, abbreviated \textbf{LAVLA}.
We investigate the latent space of a Vision-Language-Action Model, focusing on the action decoder and clustering embeddings in these layers.
Specifically, we collect model embeddings during forward passes from the two main modules of the VLA: the final layer of the Vision-Language Model (VLM) backbone and all intermediate cross-attention layers of the Diffusion Transformer (DiT)~\cite{peebles2023dit} action decoder.
Furthermore, we introduce an embedding-weighting module to amplify the features of important visuolinguistic tokens from the corresponding VLM backbone embeddings used in the cross-attention layers of the action decoder. 
We demonstrate the effectiveness of the LAVLA framework by applying it to NVIDIA's GR00T N1.5 model~\cite{nvidia2025gr00tn1openfoundation} using a subset of its original training data~\cite{openx2024}.
Finally, we extract a human-interpretable concept from the resulting clusters based on the top-$n$ closest samples to each centroid, assisted by an external VLM.
Our results underscore the pivotal role of the embedding-weighting module in guiding the action denoising process. 

Our contributions can be summarised as follows:

\begin{itemize}
    \item We propose LAVLA, a novel cluster-based framework for analysing the embeddings in VLAs and demonstrate its effectiveness by analysing the latent space of the NVIDIA GR00T N1.5 model.
    
    \item We introduce a cross-attention-based weighting module designed to amplify the influence of tokens prioritised by the action decoder. This technique performs consistent improvements across unsupervised clustering benchmarks; specifically, we observe a 62.1\% increase in Silhouette scores and an 8\% improvement in the Davies-Bouldin (DB) Index. 

    \item We include a concept extraction pipeline that assigns a human-interpretable concept to each cluster, derived from the top-$n$ nearest samples to the cluster centroid, labelled using an external VLM and evaluated using CLIP.
    
    \item Our analysis reveals that cross-attention layers consistently focus on specific tokens from the VLM backbone representations across different diffusion timesteps. The resulting clusters exhibit a clear disentanglement of spatiotemporal and kinematic features, suggesting that the cross-attention mechanism effectively incorporates the visuolinguistic context in the action diffusion process.

\end{itemize}





\begin{figure*}
    \centering
    \includegraphics[width=\textwidth]{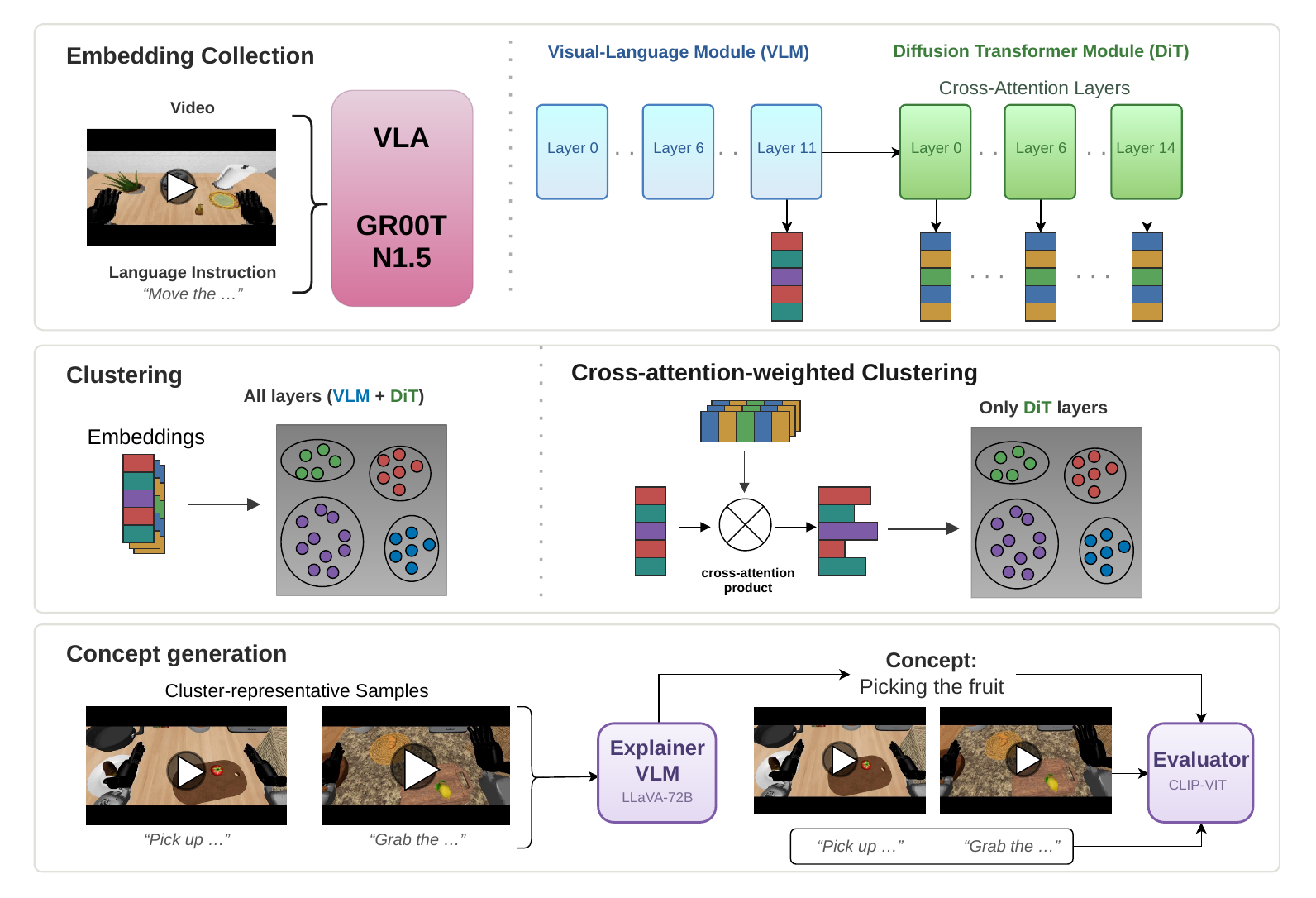}
    \caption{Overview of our framework. We collect embeddings (coloured bars) from the last VLM backbone layer and the DiT action decoder. After the collection, we compute two different cluster typologies:  baseline clustering applied to all embeddings from each layer; our embedding-weighting module steers the VLM backbone tokens according to their influence in the cross-attention layers. Finally, we generate concepts based on the top-$n$ nearest samples for each cluster and evaluate each of them with CLIP.}
    \label{fig:framework_main}
\end{figure*}
    

\section{Related Work}

\subsection{Vision-Language-Action Models}
Recent works on Vision Language Action Models~\cite{kim2025openvla,octomodelteamOctoOpenSourceGeneralist2024,blackVisionLanguageActionFlow2024,belkhaleRTHActionHierarchies2024,shiHiRobotOpenEnded2025,nvidia2025gr00tn1openfoundation} aim to create universal models capable of generating robot actions or trajectories across different embodiments. 
The training paradigms often follow those of recent LLMs: large datasets are collected that span a wide range of trajectories and scenarios to enable generalisation~\cite{openx2024}. 
Commonly, VLAs map language instructions and a visual observation to robotic actions. 
Generally, the visual input is first processed into vision token embeddings, which are then concatenated with the embedded language tokens and fed into the Large Language Model (LLM) backbone. 
The backbone LLM then produces tokens directly, which map to specific robotic actions~\cite{kim2025openvla,belkhaleRTHActionHierarchies2024}, or returns the embeddings of intermediate to final layers, which are then further processed by specific action decoder heads. 
These action decoder heads utilise different strategies to generate multi-dimensional continuous actions, including flow matching ~\cite{nvidia2025gr00tn1openfoundation,blackVisionLanguageActionFlow2024,shiHiRobotOpenEnded2025} and diffusion decoding strategies~\cite{octomodelteamOctoOpenSourceGeneralist2024}, or simple linear transformations~\cite{belkhaleRTHActionHierarchies2024}. 
Finally, to generalise to long-horizon tasks, some methods explicitly model hierarchical sub-tasks~\cite{belkhaleRTHActionHierarchies2024,shiHiRobotOpenEnded2025}.
Closest to our approach, \citet{haonMechanisticInterpretabilitySteering2025a} use clustering of FFN activations projected into token space to identify semantically meaningful directions for steering VLA behaviour. 
However, their clustering does not probe the action decoder.
We close this research gap by specifically investigating the action decoding process.

\subsection{Latent Analysis in Foundation Models}

Latent analysis methods reduce high-dimensional embeddings into more compact representations that are generally more interpretable than the original embedding, allowing them to be linked with human-interpretable concepts. 
 Sparse AutoEncoders (SAEs) are prominent approaches~\cite{cunningham_sparse_2023} that integrate an autoencoder into a layer of a pretrained model. 
SAEs contain a single hidden layer trained to maintain sparse neuron activations for each reconstructed token. 
According to the superposition theory \cite{elhage_toy_2022}, these new neurons encode the most salient information needed to rebuild the original embedding.
Although SAEs have typically been applied to LLMs, some researchers have also extended their usage to VLAs and VLMs~\cite{grant2026featurescreatedequalmechanistic,zhang_large_2024,pach_sparse_2025}.
However, SAEs require large amounts of training data, making them slow to adapt to the fast pace of robotics development.
A different line of research relies on clustering techniques to group similar embeddings. Unlike SAEs, clustering does not require large amounts of training data, making it more suitable for robotic tasks.
\citet{andeol_holistic_2023} introduced a unifying theoretical framework that reframes concept extraction as a form of dictionary learning, while \citet{hawaslyScalingDiscoveryLatent2024} evaluated the quality of discovered concepts to compare clustering algorithms. 

\section{Methodology}
A visual overview of the LAVLA framework is presented in Fig.~\ref{fig:framework_main}.
The first step in the LAVLA process is to collect the embeddings of selected layers during forward passes of a VLA. 
These embeddings are then clustered layer-wise using an unsupervised clustering algorithm. 
Optionally, a cross-attention-based embedding-weighting module and a PCA-based dimensionality reduction can be applied before the clustering; when both are utilised, embedding-weighting precedes dimensionality reduction.
Finally, for each cluster, we select the top-$n$ representative samples, generate a concept label using an external VLM, and evaluate it using CLIPSIM~\cite{liu2023evalcrafter}.

\subsection{Embedding Collection}
\label{sec:embedding_collection}

Our investigation of the inner workings of VLAs focuses on NVIDIA's GR00T N1.5 model~\cite{nvidia2025gr00tn1openfoundation}, a state-of-the-art, 3B parameter VLA.
GR00T N1.5 consists of an Eagle VLM~\cite{li2025eagle2} backbone to process vision and language inputs into multimodal representations and an action decoder head, which produces the actual output command.
The action decoder head is a diffusion transformer module that alternates between self-attention and cross-attention blocks, which denoise trajectories into continuous action vectors.
The action decoder head receives noisy action tokens and the encoded robot state as input, while the fused vision-language embeddings from the Eagle backbone are exclusively integrated in the cross-attention layers.
For each sample in our data selection, we collect the embeddings resulting from each cross-attention layer in the action decoder head of GR00T N1.5 as well as the 12th layer of the Eagle backbone, the last VLM layer before visuolingual embeddings are passed to the action decoder.
The action decoder performs four iterative denoising steps; consequently, each layer is traversed four times to progressively refine the initially noisy action tokens.
We store the embeddings of the action decoder with respect to the corresponding diffusion timestep.
We use a part of the Open-X-Embodiment dataset\footnote{The data is publicly available at: \href{https://huggingface.co/datasets/NVIDIA/PhysicalAI-Robotics-GR00T-X-Embodiment-Sim}{https://huggingface.co/datasets/NVIDIA/PhysicalAI-Robotics-GR00T-X-Embodiment-Sim}} for our experiments. 
More details on the data preparation can be found in Appendix~\ref{app:data}.

\subsection{Clustering}
Once the embeddings have been collected, we cluster the embeddings of all cross-attention layers of the diffusion transformer action decoder head and the last layer of the GR00T Vision-Language-Action Model using a hierarchical clustering method.
Specifically, we use agglomerative hierarchical clustering. 
Initially, we explored DBSCAN and k-means but ultimately selected agglomerative clustering based on its superior early-stage performance.
The algorithm initialises by assigning each sample to its own cluster, then recursively merges the closest clusters based on a predefined distance threshold and linkage method.
This process continues until either no more merges are possible or an alternate stopping criterion is satisfied. 
We perform agglomerative clustering until no further merges are possible.

\subsubsection{Cross-Attention-Based Embedding-Weighting}
\label{sec:clustering-cross-attention}
Early in the action denoising process of the VLA diffusion head, the embeddings contain mostly noise. 
Only in the later layers and after several diffusion steps, the model generates the final output from a more defined latent space.
In the case of the GR00T model, cross-attention layers are used during the denoising process, which incorporate the visuolingual outputs of the backbone VLM at each timestep.
We want to measure the focus allocated to each visuolingual source token $S_{VL}$ during the forward pass in GR00T's action diffusion head, to gain insight into what matters to the cross-attention layer at a specific time and reduce the impact of less impactful tokens. 
This also eliminates the disturbance of the clustering due to a noisy latent space early in the denoising process.\\
We extract the attention weight matrix $\alpha$ directly from the cross-attention layers. 
This matrix has dimensions $a \times n$, with $a$ referring to the length of the action token sequence and $n$ referring to the length of the visuolanguage token sequence. 
In the case of cross-attention in GR00T's action decoder, the query matrix $Q_A$ originates from encoding the previous self-attention layer, while the key matrix $K_{VL}$ is projected from the injected vision-language backbone embeddings. 
Following standard practice in scaled dot-product attention, the scores are scaled by the factor $\sqrt{d_k}$ to mitigate softmax saturation.
\begin{equation}
     \alpha^{a \times n} = \text{softmax}\left(\frac{Q_AK_{VL}^T}{\sqrt{d_k}}\right) 
\end{equation}
We average the weights in $\alpha$ for each source token $j$ across all target tokens $i$ to receive a vector of impact scores $M = [m_1, \dots, m_{n}]$, with $m_j$ defined as:
\begin{equation}
     m_j = \frac{1}{n} \sum_{i=1}^{n} \alpha_{ij} 
\end{equation}
Next, we reduce across all heads in multi-head attention by using the mean before applying the normalisation. 
We receive the normalised weight per source token by applying the softmax function to the vector of impact scores $M$. 
This results in the normalised weight vector $W = [w_1, \dots, w_{n}]$, with:
\begin{equation}
    w_j = \frac{\exp(m_j)}{\sum_{k=1}^{n} \exp(m_k)}
\end{equation}
Finally, we apply the Hadamard product $\odot$ between the normalised weight vector $W$ and the visuolingual tokens $S'_{VL} = S_{VL} \odot W$.
This reduces the impact of less influential tokens while boosting the more important ones in the resulting weighted vision-language token sequence $ S'_{VL}$. The embeddings $S'_{VL}$ are specific to each cross-attention layer and diffusion timestep, allowing us to analyse the importance of tokens in the backbone vision-language sequence during the action denoising process.

\subsubsection{Cluster Evaluation}
To evaluate layer-wise cluster quality, we assess the spatial density and separation of the identified clusters using three standard metrics: the Silhouette Coefficient~\cite{silhouette}, the Caliński-Harabasz (CH) Index~\cite{chindex}, and the Davies-Bouldin (DB) Index~\cite{dbindex}. 
These metrics are applied across all layers of interest as defined in Section~\ref{sec:embedding_collection}.
Higher CH Index, lower DB Index, and higher Silhouette score indicate better separation of clusters.\\
To further investigate the flow of individual samples through the network, we calculate the Overlap Coefficient (OC)~\cite{vermaComparativeAnalysisSimilarity2020} and Jaccard Index~\cite{vermaComparativeAnalysisSimilarity2020} between all possible pairs of clusters $A$ and $B$. 
See Appendix \ref{appendix:formulas} for the exact formulations.  
The Overlap Coefficient highlights containment of the smaller set within the larger, and reaches its maximum value of 1 when the smaller set is perfectly contained within the larger set.
On the other hand, the Jaccard index balances the amount of shared elements and elements only contained in one of the two sets. 
The maximum value of 1 is only reached if both sets are identical.
To compare cluster assignments of different experiments and layers, we perform an additional alignment step as detailed in Appendix~\ref{app:alignment}.

\subsection{Concept Generation}
\label{sec:concept-generation}
After completing the cluster computation, we generate a representative concept for each cluster. Based on prior work~\cite{zhang_large_2024,liu2023evalcrafter}, we select the top-$n$ cluster samples nearest to the centroid and query a VLM to identify the human-interpretable pattern shared across them. 
Each sample consists of two elements: the textual command and the action video, which is decomposed into a sequence of frames, 10 total frames for a 5-second clip. The VLM is then prompted to generate a short sentence that summarises each sample separately. 
These resulting sentences are further passed together to the VLM a second time to synthesise a single concept label for the cluster.
Finally, we evaluate the overall performance using the CLIPSIM metric~\cite{liu2023evalcrafter}, computed as the CLIP similarity between the final concept and the single frames of each cluster sample.

\section{Experiment Setup}
Embeddings for our analysis were collected on a single NVIDIA A100 GPU using a fully frozen GR00T N1.5 model. 
We collect 1,000 embeddings per task from the 24 tasks in the dataset in ca. 4 hours.
Consecutive samples of the same task and episode in the resulting 24,000 embeddings are offset by 10 frames.
We cluster the embeddings using agglomerative clustering with Euclidean distance as metric and average linkage.
The latent representations change between individual layers, hence there is no one-size-fits-all distance measure for all layers.
We notice, however, that it yields the best results within certain ranges within the same components. 
We present an exhaustive list of hyperparameters in Appendix~\ref{app:hyperparameters}. 
For the concept generation part, we adopt LLaVA 72B~\cite{liu2024llavanext} as the generator, and set top-$n$=5 due to computational constraints. For the same reason, we limited our analysis to the 10 most populous clusters for each layer/cluster technique pair. This threshold is specific to non-PCA-cluster methods, since these techniques create, on average, 7 clusters, as shown in Table~\ref{tab:clustering_metrics}. The concept generation pipeline was run on two NVIDIA A100 GPUs for ca. 3 hours.

\section{Results}
\begin{figure}[ht]
    \centering
    \begin{subfigure}{0.45\textwidth}
        \centering
        \includegraphics[trim={0 0.7cm 0 0}, clip, width=\linewidth]{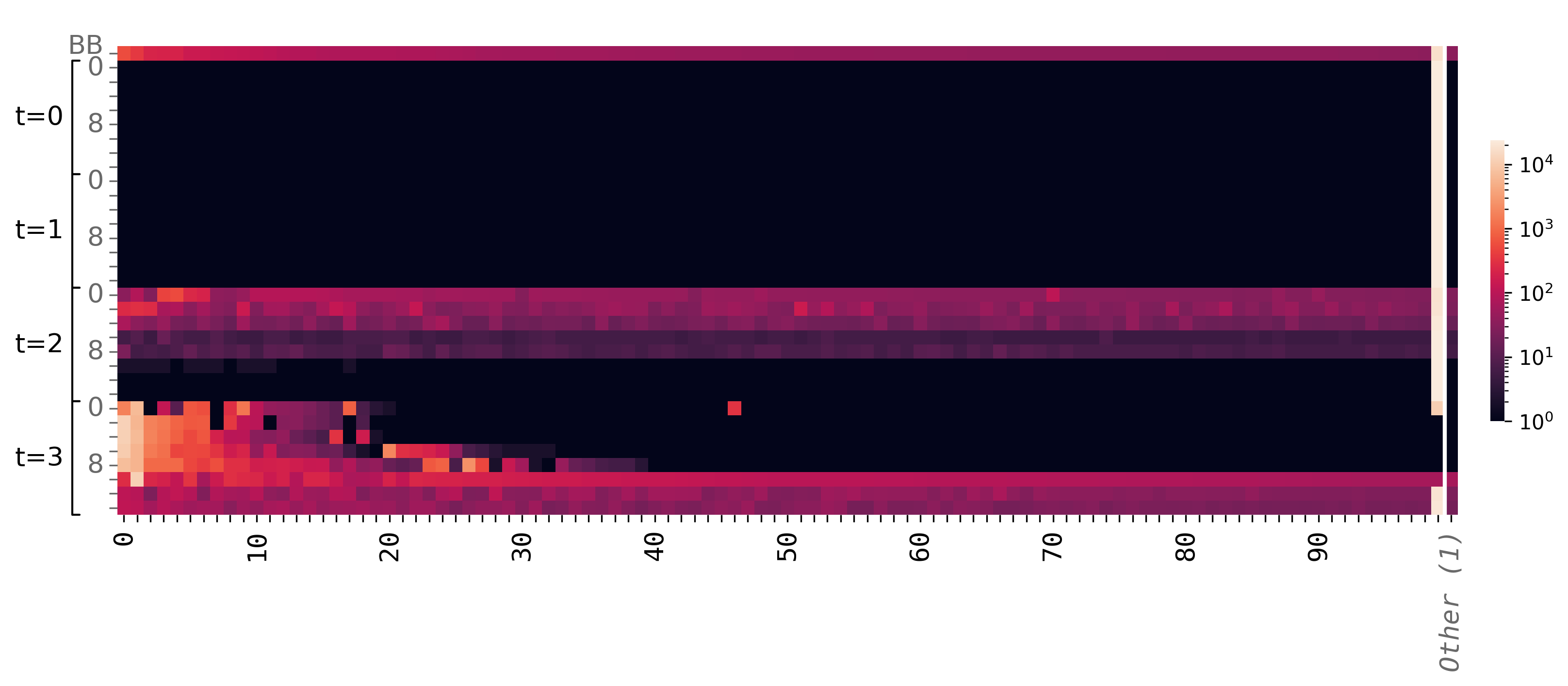}
        \caption{Baseline clusters through layers}
        \label{fig:cluster_distribution_unweighted}
    \end{subfigure}
    \hfill
    \begin{subfigure}{0.45\textwidth}
        \centering
        \includegraphics[trim={0 0.7cm 0 0}, clip, width=\linewidth]{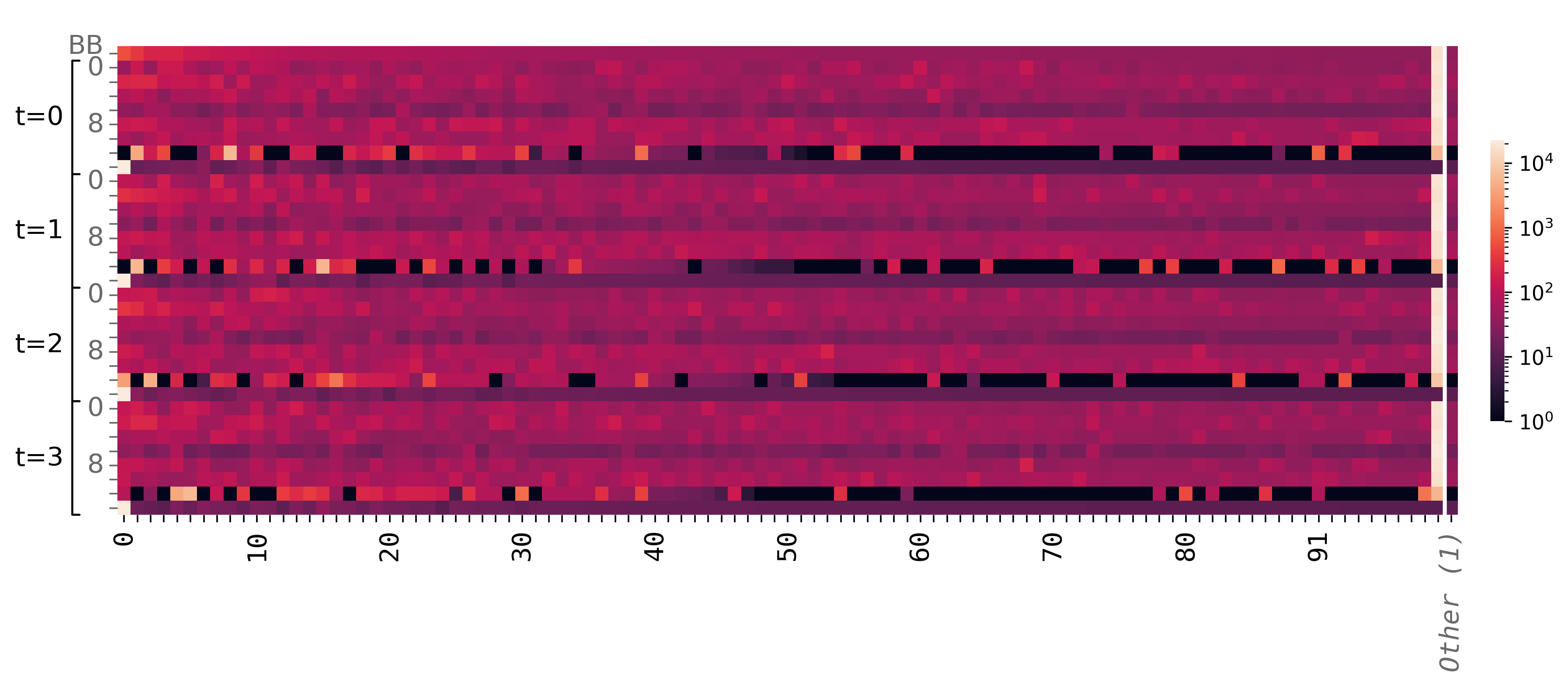}
        \caption{Weighted Clustering clusters through layers}
        \label{fig:cluster_distribution_weighted}
    \end{subfigure}
    \hfill
    \begin{subfigure}{0.45\textwidth}
        \centering
        \includegraphics[trim={0 0.7cm 0 0}, clip, width=\linewidth]{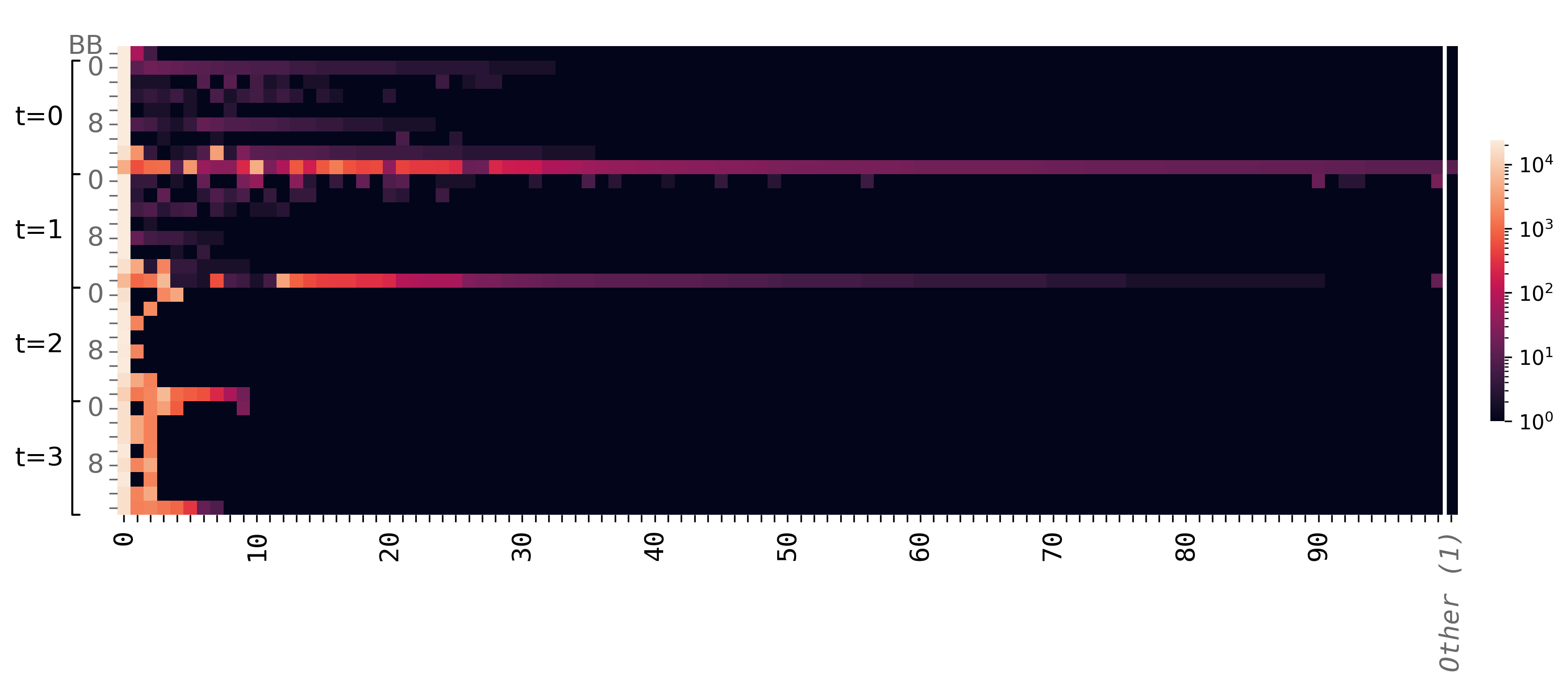}
        \caption{PCA Clustering clusters through layers}
        \label{fig:cluster_distribution_pca}   
    \end{subfigure}
    \hfill
    \begin{subfigure}{0.45\textwidth}
        \centering
        \includegraphics[trim={0 0.7cm 0 0}, clip, width=\linewidth]{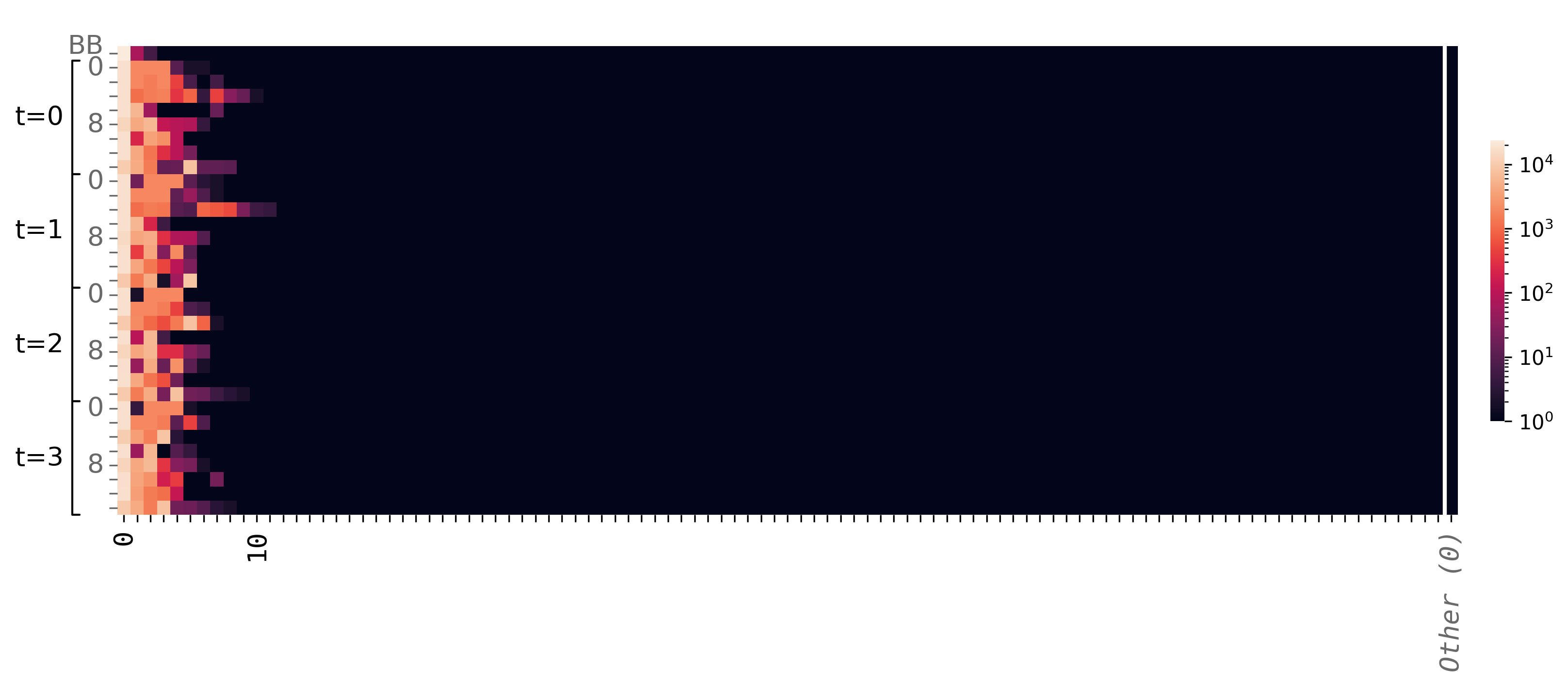}
        \caption{Weighted PCA Clustering clusters through layers}        \label{fig:cluster_distribution_weighted_pca}   
    \end{subfigure}
    \caption{Heatmaps of cluster sizes across different layers and LAVLA configurations.}
\end{figure}

To evaluate the effectiveness of the LAVLA framework and investigate the latent space of the GR00T action decoding process, we investigate how clusters evolve during the diffusion forward passes, highlight patterns visible in the cluster contents, and present quantitative metrics on the effectiveness of different LAVLA configurations. We present our results across individual layers. 
Cross-attention layer weights are reused during the 4-step diffusion process (beginning with step 0, ending at step 3). 
Accordingly, we further differentiate embeddings from cross-attention layers by their diffusion timestep t, i.e. the cross-attention layer 2 with t=1 and cross-attention layer 2 with t=3 use the same weights, but the embeddings have been extracted at diffusion timestep t=1 and t=3 respectively. 
Results referencing the embeddings of the backbone layer are annotated as "BB", while we provide the layer index (starting at 0) to differentiate cross-attention layers (for example, see the y-axis of Figure~\ref{fig:cluster_distribution_unweighted}). 

\subsection{Clustering Results}
\textbf{Cross-attention-based embedding-weighting leads to measurable gains in clustering performance.}
We compare the general performance of different LAVLA configurations using common unsupervised cluster-evaluation metrics, and display the detailed metrics in Table~\ref{tab:clustering_metrics}.
Adding the embedding-weighting mechanism has an overall positive effect, especially in the case of unreduced embeddings.
It improves the Silhouette scores by 62.1\% in the scenario without PCA and by 24.0\% in the scenario with PCA. 
DB Index scores improve in each scenario by ca. 8\%. 
This is likely due to the embedding-weighting mechanism not relying on noisy action tokens but rather on differently amplified backbone embeddings, which bear semantic meaning over the full process. 
While the action token embeddings start as noise, they only become semantically informative towards the end of the diffusion process.
CH Index scores only improve in the case in which PCA is also applied. 
This is likely a result of the extremely high number of clusters in the unweighted case, which leads to dense clusters with only a single sample, inflating the resulting score.
Based on these results, we hypothesise that the embedding-weighting mechanism would also positively impact other interpretability frameworks that investigate model latent spaces, e.g., SAEs, and suggest their inclusion in future work.  
\begin{table*}[!ht]
\centering
\caption{Clustering performance across configurations}
\small
\begin{tabular}{lcccc}
\toprule
\textbf{LAVLA} & \textbf{Silhouette Score}$\uparrow$ & \textbf{DB Index}$\downarrow$ & \textbf{CH Index}$\uparrow$ & \textbf{N clusters} \\
\midrule
Baseline & 0.0837 $\pm$ 0.05 & 1.4734 $\pm$ 0.18 & 402.12 $\pm$ 358.6 & 15703.8 $\pm$ 10282.3 \\
w/ weighted & 0.1357 $\pm$ 0.03 & \textbf{1.3617 $\pm$ 0.27} & 56.41 $\pm$ 21.08 & 3835.2 $\pm$ 3479.0 \\
w/ PCA & 0.1324 $\pm$ 0.06 & 1.5858 $\pm$ 0.24 & 1063.9 $\pm$ 481.4 & 25.93 $\pm$ 72.1 \\
 w/ PCA w/ weighted &\textbf{0.1642 $\pm$ 0.06} & 1.4682 $\pm$ 0.15 & \textbf{1651.0 $\pm$ 775.8} & 7.1515 $\pm$ 2.41 \\
\bottomrule
\end{tabular}
\label{tab:clustering_metrics}
\end{table*}

\begin{figure}[ht]
    \centering
    \includegraphics[width=\linewidth]{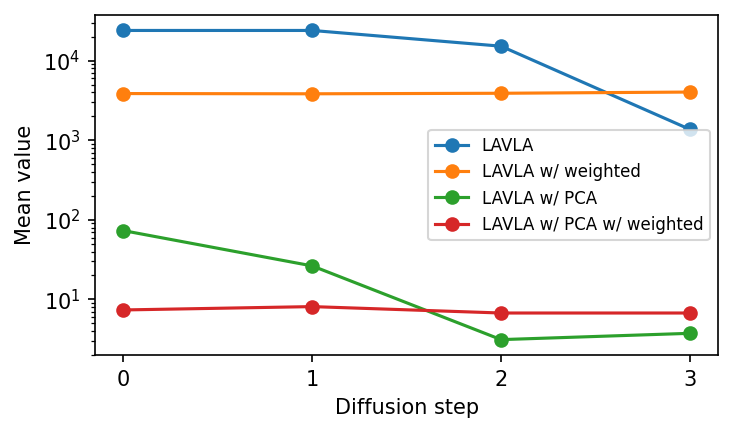}
    \caption{Mean number of clusters over diffusion timesteps}
    \label{fig:n_clusters}
\end{figure}
\textbf{Latent representations exhibit increased structural coherence at deeper architectural layers and later diffusion timesteps.}
The unreduced latent embeddings of the GR00T model are represented as a $(n \times d)$ matrix, with $n$ referring to the sequence length of the vision and language input tokens, and $d$ being the internal dimension of the per-token embeddings. 
The vision-language token sequence length varies slightly with the number of tokens in the language instruction; the number of extracted vision tokens remains constant.
Averaging across the sequence length results in a set of $d$-dimensional embedding vectors that may have been amplified by the embedding-weighting process depending on the experiment configuration. 
Influenced by the curse of dimensionality~\cite{curseOfDimensionality}, applying agglomerative clustering on these high-dimensional sets of embedding vectors results in a large number of clusters, with only a few select clusters containing more than a few samples. 
This is highlighted by the heatmaps in Figures~\ref{fig:cluster_distribution_unweighted} to~\ref{fig:cluster_distribution_weighted_pca} which display the largest 100 clusters over layers and diffusion timesteps.

As shown in Figure~\ref{fig:n_clusters}, the latent representations become increasingly structured as the diffusion process nears completion.
This progression toward a more organised state leads to a lower cluster count, suggesting that disparate embeddings are being consolidated into cohesive groups.
Applying PCA for dimensionality reduction reduces the embedding dimension to 100 per embedding. 
Consequently, the overall number of extracted clusters drops significantly. 
We observe the same trend as with the uncompressed embeddings, where the number of clusters becomes lower towards later diffusion timesteps. 
A visual inspection of the latent space under PCA visualisations reveals the same organising pattern in Appendix~\ref{app:pca}.

However, this same trend is less prominent when incorporating the embedding-weighting mechanism. 
This likely stems from the weighted backbone embeddings being directed towards certain clusters during the cross-attention mechanism throughout the whole process. 
In the absence of this weighting mechanism, the noisy action embeddings require additional denoising iterations before the action decoder layers can resolve the latent space into a structured state.
The UMAP visualisations in Appendix~\ref{app:umap} further highlight the effect of the weighting mechanism.


\subsection{Cluster Consistency}
\begin{figure}[ht]
    \centering
    \begin{subfigure}{0.45\linewidth}
    \centering
        \includegraphics[trim={0 0 0 1.05cm}, clip, width=\linewidth]{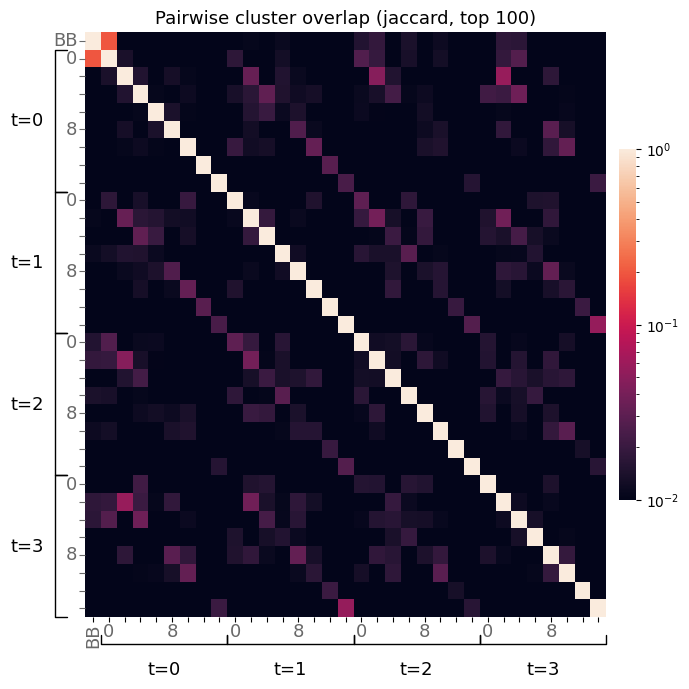}
        \caption{Weighted Jaccard}
        \label{fig:weighted_overlap_jaccard}
    \end{subfigure}
    \hspace{1em}
    \begin{subfigure}{0.45\linewidth}
    \centering
        \includegraphics[trim={0 0 0 1.05cm}, clip, width=\linewidth]{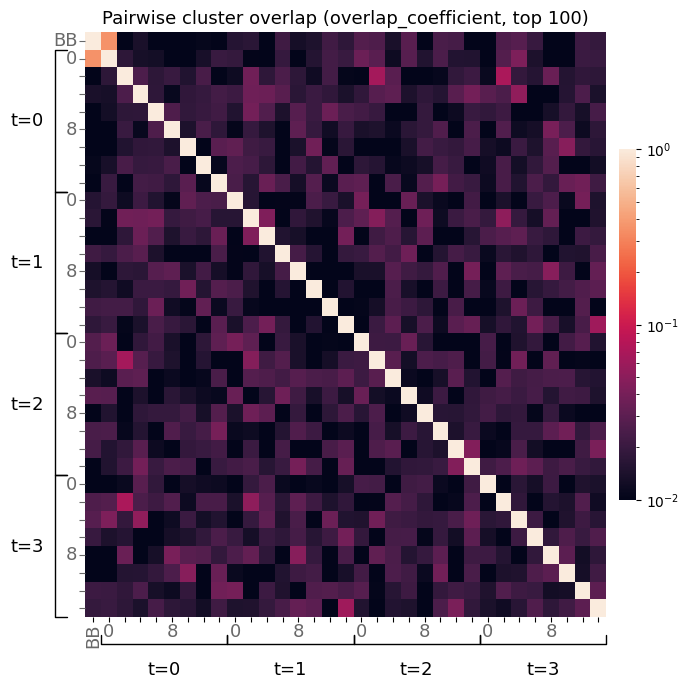}
        \caption{Weighted OC}
        \label{fig:weighted_overlap_oc}
    \end{subfigure}
    
    \begin{subfigure}{0.45\linewidth}
    \centering
        \includegraphics[trim={0 0 0 1.05cm}, clip, width=\linewidth]{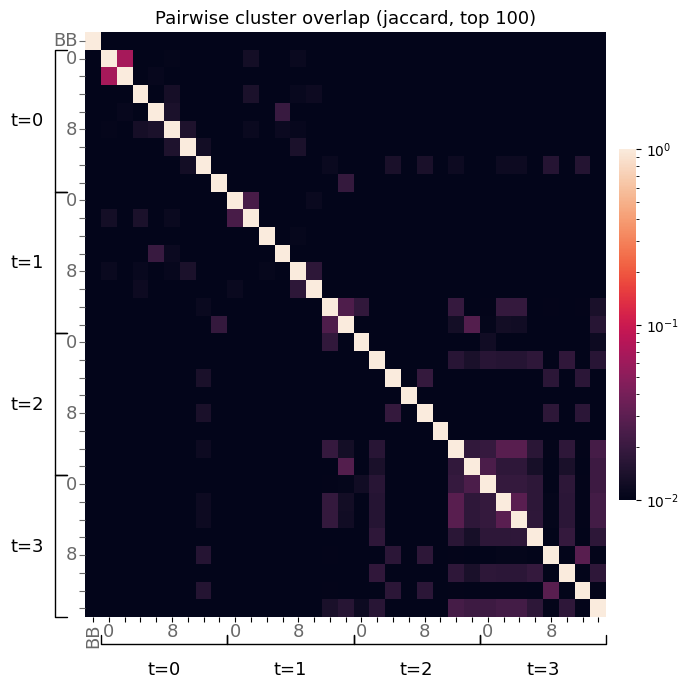}
        \caption{PCA Jaccard}
        \label{fig:pca_overlap_jaccard}
    \end{subfigure}
    \hspace{1em}
    \begin{subfigure}{0.45\linewidth}
    \centering
        \includegraphics[trim={0 0 0 1.05cm}, clip, width=\linewidth]{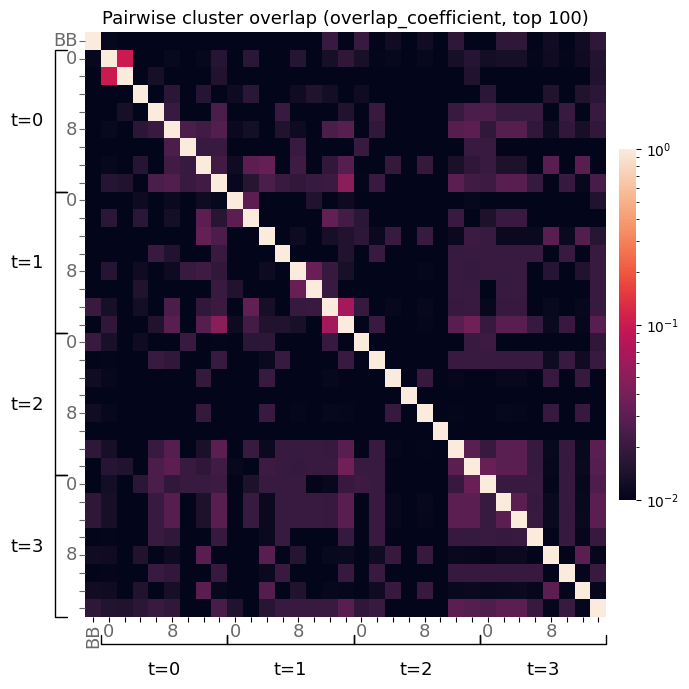}
        \caption{PCA OC}
        \label{fig:pca_overlap_oc}
    \end{subfigure}
    
    \begin{subfigure}{0.45\linewidth}
    \centering
        \includegraphics[trim={0 0 0 1.05cm}, clip, width=\linewidth]{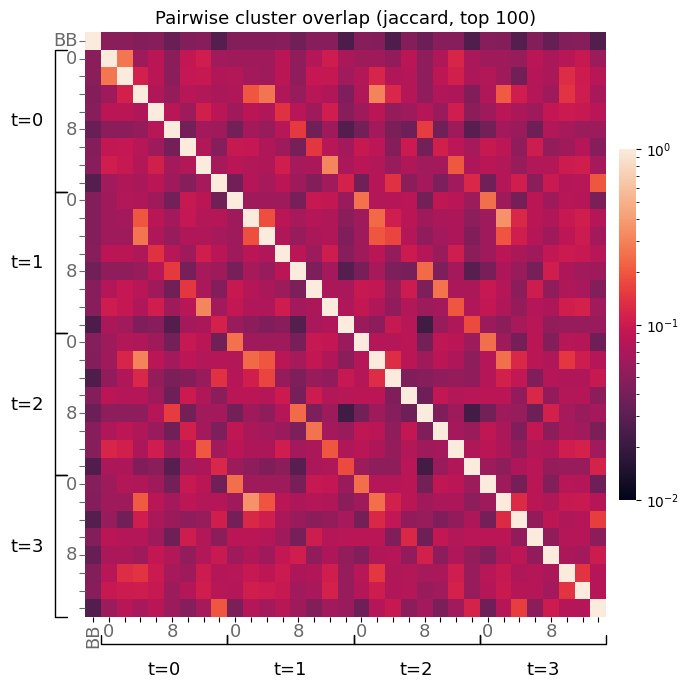}
        \caption{Weighted + PCA Jaccard}
        \label{fig:weighted_pca_overlap_jaccard}
    \end{subfigure}
    \hspace{1em}
    \begin{subfigure}{0.45\linewidth}
    \centering
        \includegraphics[trim={0 0 0 1.05cm}, clip, width=\linewidth]{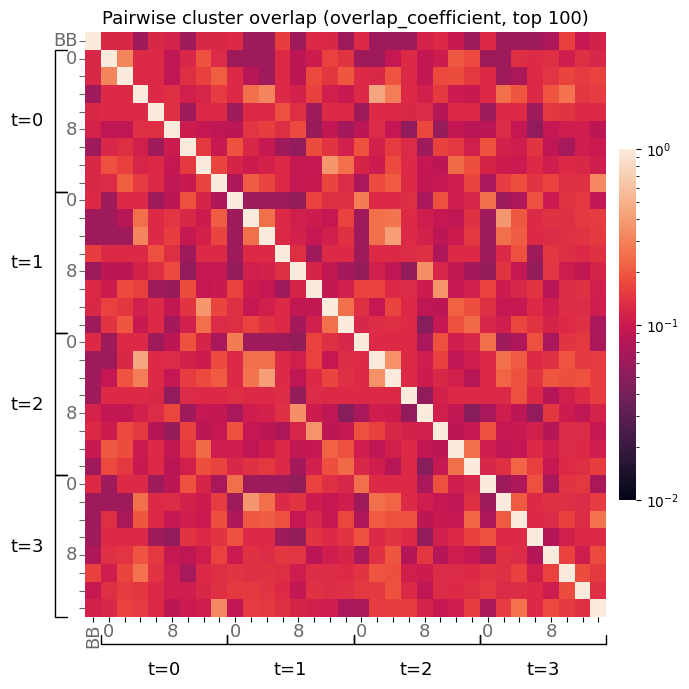}
        \caption{Weighted + PCA OC}
        \label{fig:weighted_pca_overlap_oc}
    \end{subfigure} 
    \caption{Layer-wise cluster overlap matrices for different LAVLA configurations.}
\end{figure}

\textbf{Cluster assignments are stable across diffusion timesteps within the same layer.}
To investigate whether the cluster assignments remain consistent across different layers, we computed the pairwise Jaccard and Overlap Coefficient metrics of the largest 100 clusters between layers and diffusion steps. 
The results are displayed in the heatmaps in Figures~\ref{fig:weighted_overlap_oc} to ~\ref{fig:weighted_pca_overlap_oc}.
We did not include the heatmaps for the case without either PCA or the embedding-weighting mechanism, as the cluster algorithm was unable to find meaningful assignments during early diffusion timesteps (see Figure~\ref{fig:n_clusters}) and assigns each sample to its own cluster. 
As a result, this inflates the corresponding metrics in the early layers.
The heatmaps~\ref{fig:pca_overlap_jaccard} and~\ref{fig:pca_overlap_oc} reveal mainly two things: first, cluster assignments stay more consistent closer to the final outputs. 
This is largely a consequence of the diffusion process, which leads to less noisy embeddings over time, and the model having to decide on a much smaller number of possible action vectors compared to the vast possibilities of language-vision-state inputs. 
Second, the repeating patterns in the heatmaps suggest that cluster assignments between the same layer but different diffusion timesteps remain consistent. 
They are especially highlighted by the diagonal lines visible in Figures~\ref{fig:weighted_overlap_oc},~\ref{fig:weighted_pca_overlap_jaccard}, and~\ref{fig:weighted_pca_overlap_oc}.
We hypothesise that this is due to the model using different layers to pay special attention to certain features in the backbone embeddings. 
E.g., one layer might consider colour representations more important while another might specifically incorporate object-centric features. 
Figures~\ref{fig:pca_overlap_jaccard} and~\ref{fig:pca_overlap_oc} further illustrate how the latent space becomes more ordered towards the end of the diffusion process, likely resulting in more consistent clusters (see patterns in bottom right corner).
Our LAVLA framework cannot reveal which features precisely are important for different layers, and as such we leave this investigation for future work on VLAs.
\begin{figure*}[!ht]
    \centering
    \begin{subfigure}{0.24\textwidth}
        \centering
        \includegraphics[width=\linewidth]{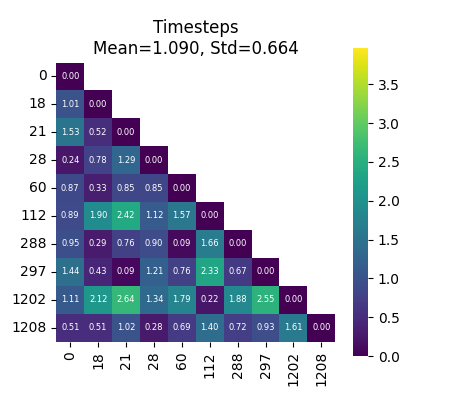}
        \caption{Timesteps}
        \label{fig:bb_timesteps}
    \end{subfigure}
    \hfill
    \begin{subfigure}{0.24\textwidth}
        \centering
        \includegraphics[width=\linewidth]{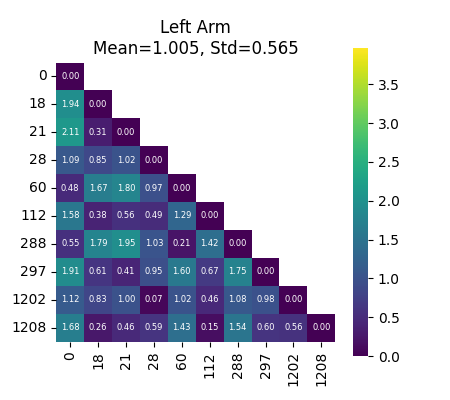}
        \caption{Left arm displacement}
        \label{fig:bb_left_arm}
    \end{subfigure}
    \hfill
    \begin{subfigure}{0.24\textwidth}
        \centering
        \includegraphics[width=\linewidth]{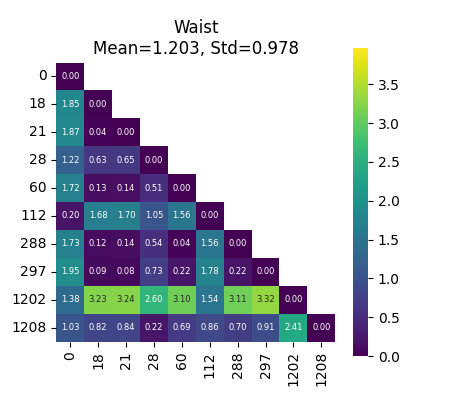}
        \caption{Waist displacement}
        \label{fig:bb_waist}
    \end{subfigure}
    \hfill
    \begin{subfigure}{0.24\textwidth}
        \centering
        \includegraphics[width=\linewidth]{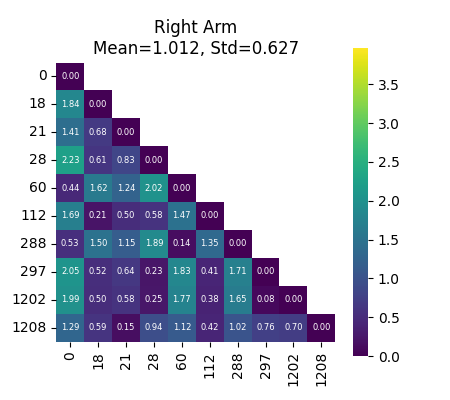}
        \caption{Right arm displacement}
        \label{fig:bb_right_arm}
    \end{subfigure}
    \caption{Wasserstein distances between feature distributions of top clusters within VL backbone}
    \label{fig:wasser_bb}
\end{figure*}
    
\begin{figure*}[!ht]
    \centering
    \begin{subfigure}{0.24\textwidth}
        \centering
        \includegraphics[width=\linewidth]{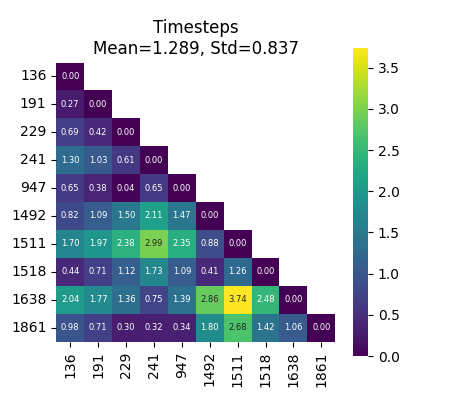}
        \caption{Timesteps}
        \label{fig:ca14_timesteps}
    \end{subfigure}
    \hfill
    \begin{subfigure}{0.24\textwidth}
        \centering
        \includegraphics[width=\linewidth]{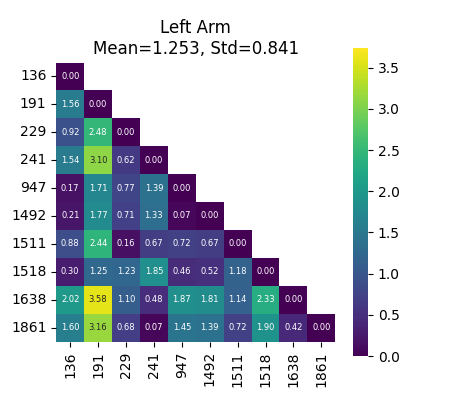}
        \caption{Left arm displacement}
        \label{fig:C14dt3W_left_arm}
    \end{subfigure}
    \hfill
    \begin{subfigure}{0.24\textwidth}
        \centering
        \includegraphics[width=\linewidth]{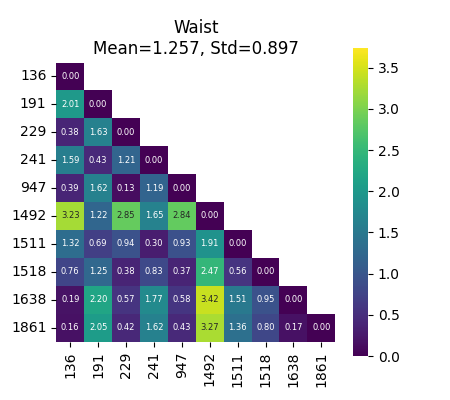}
        \caption{Waist displacement}
        \label{fig:C14dt3W_waist}
    \end{subfigure}
    \hfill
    \begin{subfigure}{0.24\textwidth}
        \centering
        \includegraphics[width=\linewidth]{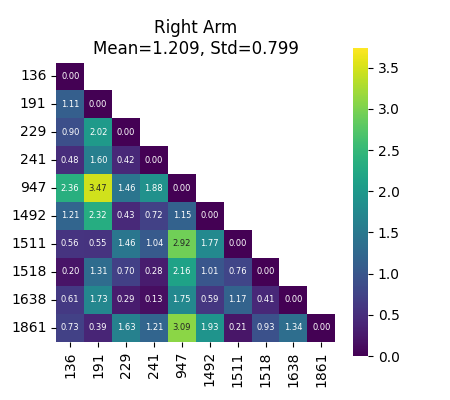}
        \caption{Right arm displacement}
        \label{fig:C14dt3W_right_arm}
    \end{subfigure}
    \caption{Wasserstein distances between feature distributions of top clusters of cross-attention layer 14 at final diffusion timestep (t=3).}
    \label{fig:wasser_ca14}
\end{figure*}
\subsection{Latent Space Organisation}
\textbf{Clusters exhibit distinct temporal characteristics.}
As shown in Figures~\ref{fig:bb_timesteps} and~\ref{fig:ca14_timesteps}, the latent space of the action diffusion head exhibits higher temporal separability than the VLM backbone.
Besides a visible increase in separability between clusters, this observation is supported by the increased mean within the displayed 10 largest clusters.
This suggests that the model discriminates between generalisable states commonly encountered at specific intervals within different episodes, implying spatiotemporal awareness. 
Furthermore, even if this generalisation were driven solely by visual features common to specific episode timesteps, it would nonetheless indicate that the model has developed a structured spatial representation of the task environment.
Further experiments are required to fully comprehend the temporal understanding capabilities of the model.

\textbf{Clusters separate body part usage.}
The plots in Figures~\ref{fig:bb_left_arm} to~\ref{fig:bb_right_arm}, and~\ref{fig:C14dt3W_left_arm} to~\ref{fig:C14dt3W_right_arm} reveal the same trend with a ca. 20\% increase in mean intra-cluster Wasserstein distance for the arm joints observed in Figures~\ref{fig:bb_left_arm} and~\ref{fig:C14dt3W_left_arm} for the left arm, and Figures~\ref{fig:bb_right_arm} and~\ref{fig:C14dt3W_right_arm} for the right arm. 
While less prominent, it can still be observed in the waist embeddings in Figures~\ref{fig:bb_waist} and~\ref{fig:C14dt3W_waist}.
\begin{table*}[!ht]
\caption{Mean $\pm$ std CLIPSIM$\uparrow$ (video vs.\ concept) per configuration and layer.}
\label{tab:clipsim_all_layers}
\centering
\small
\begin{tabular}{lccccccccc}
\toprule
Layer & Backbone L11 &  CA0 &  CA6 &  CA12 &  CA14 \\
\midrule
unweighted & 0.214 $\pm$ 0.014 & \textbf{0.229} $\pm$ 0.009 & 0.228 $\pm$ 0.011 & 0.226 $\pm$ 0.014 & 0.218 $\pm$ 0.016 \\
weighted & 0.214 $\pm$ 0.014 & 0.225 $\pm$ 0.016 & 0.216 $\pm$ 0.015 & 0.206 $\pm$ 0.019 & 0.213 $\pm$ 0.024 \\
pca & \textbf{0.219} $\pm$ 0.013 & 0.223 $\pm$ 0.015 & \textbf{0.233} $\pm$ 0.018 & \textbf{0.233} $\pm$ 0.022 & \textbf{0.221} $\pm$ 0.020 \\
weighted\_pca & \textbf{0.219} $\pm$ 0.013 & 0.223 $\pm$ 0.015 & 0.209 $\pm$ 0.024 & 0.225 $\pm$ 0.010 & \textbf{0.221} $\pm$ 0.014 \\
\bottomrule
\end{tabular}
\end{table*}

\subsection{Concept Generation Results}

Table~\ref{tab:clipsim_all_layers} reports CLIPSIM results across the four main configurations. 
PCA-based and weighted variants tend to yield slightly more faithful concepts, with PCA consistently improving performance across layers. 
However, CLIPSIM scores should be interpreted with caution: robotic manipulation datasets typically exhibit low visual diversity and the input clips are short, which likely imposes a ceiling on achievable performance regardless of the clustering method. 
For the same reason, the final concepts identified are semantically close to each other, mostly involving picking objects and returning them into a container. Nevertheless, we believe that since the DiT is a denoising process, the concepts may persist or evolve across layers rather than remain fixed, while still focusing on the same macro task.
This experiment demonstrates the feasibility of concept construction, though concept quality remains a limitation to address in future work. 
We present some generated concept examples in Appendix~\ref{app:concept_generation}. 


\section{Conclusion}
We present LAVLA, a framework for analysing the latent spaces of Vision-Language-Action (VLA) Models through clustering.
By amplifying the tokens prioritised during cross-attention, LAVLA elucidates the connection between backbone feature extraction and the iterative denoising process of the action decoder.
Through comparative experiments, we demonstrate that our embedding-weighting mechanism significantly enhances representational clarity.
Our results indicate that task-specific representations are distributed across distinct cross-attention layers and maintain high spatiotemporal consistency across the diffusion trajectory.
By making VLA latent spaces more transparent, LAVLA moves us closer to trustworthy, language-driven robotic systems.


\section*{Ethical Considerations}
The presented work aims at improving the understanding of the inner workings of VLAs. 
The LAVLA framework poses no inherent risk to aspects such as data privacy, as models and data used in this work are publicly available. 
Investigating the latent space of pretrained models also does not introduce risks regarding inclusivity and biases, but helps in detecting these.
Deploying VLAs in real-world scenarios on a variety of embodiments could lead to catastrophic consequences if the model is not tested and thoroughly understood. 
With increased understanding of the VLA latent space, LAVLA improves the safety of VLAs.

\section*{Limitations}
Our experiments are limited to a single dataset and model; results may not generalise to other datasets or architectures, which we leave for future work. Concept generation and its unsupervised semantic verification remain challenging without ground-truth annotations, and since concepts are derived from static images, temporal or dynamic aspects of scenes are more difficult to capture. 
Our analysis is bounded by the VLA's input horizon, which only covers short clips, so our findings may not extend to longer-horizon behaviours.
Finally, our clustering analysis relies on agglomerative clustering, whose assumptions may not fully capture the structure of complex, non-spherical manifolds, potentially influencing the resulting concept groupings.


\bibliography{references}

@InProceedings{kim2025openvla,
  title = 	 {{OpenVLA: An Open-Source Vision-Language-Action Model}},
  author =       {Kim, Moo Jin and Pertsch, Karl and Karamcheti, Siddharth and Xiao, Ted and Balakrishna, Ashwin and Nair, Suraj and Rafailov, Rafael and Foster, Ethan P and Sanketi, Pannag R and Vuong, Quan and Kollar, Thomas and Burchfiel, Benjamin and Tedrake, Russ and Sadigh, Dorsa and Levine, Sergey and Liang, Percy and Finn, Chelsea},
  booktitle = 	 {Proceedings of The 8th Conference on Robot Learning},
  pages = 	 {2679--2713},
  year = 	 {2025},
  editor = 	 {Agrawal, Pulkit and Kroemer, Oliver and Burgard, Wolfram},
  volume = 	 {270},
  series = 	 {Proceedings of Machine Learning Research},
  month = 	 {11},
  publisher =    {PMLR},
  address = {Munich, Germany}
}

@article{zang2025pretrainedvisionlanguagemodelslearn,
title={{Pre-trained Vision-Language Models Learn Discoverable Visual Concepts}},
author={Yuan Zang and Tian Yun and Hao Tan and Trung Bui and Chen Sun},
journal={Transactions on Machine Learning Research},
volume = {2025},
issn={2835-8856},
year={2025},
url={https://openreview.net/forum?id=Vq0wMFBjo2},
numpages={22}
}

@inproceedings{curseOfDimensionality,
  title = {When {{Is}} ``{{Nearest Neighbor}}'' {{Meaningful}}?},
  booktitle = {Database {{Theory}} --- {{ICDT}}'99},
  author = {Beyer, Kevin and Goldstein, Jonathan and Ramakrishnan, Raghu and Shaft, Uri},
  editor = {Beeri, Catriel and Buneman, Peter},
  year = 1999,
  pages = {217--235},
  publisher = {Springer},
  address = {Berlin, Heidelberg},
  doi = {10.1007/3-540-49257-7_15},
  isbn = {978-3-540-49257-3},
  langid = {english}
}

@article{vermaComparativeAnalysisSimilarity2020,
  title = {A Comparative Analysis of Similarity Measures Akin to the {{Jaccard}} Index in Collaborative Recommendations: Empirical and Theoretical Perspective},
  shorttitle = {A Comparative Analysis of Similarity Measures Akin to the {{Jaccard}} Index in Collaborative Recommendations},
  author = {Verma, Vijay and Aggarwal, Rajesh Kumar},
  year = 2020,
  month = jun,
  journal = {Social Network Analysis and Mining},
  volume = {10},
  number = {1},
  eid = {43},
  issn = {1869-5469},
  doi = {10.1007/s13278-020-00660-9},
  urldate = {2026-04-23},
  langid = {english}
}

@article{chindex,
author = {T. Caliński and J Harabasz},
title = {{A dendrite method for cluster analysis}},
journal = {Communications in Statistics},
volume = {3},
number = {1},
pages = {1--27},
year = {1974},
publisher = {Taylor \& Francis},
doi = {10.1080/03610927408827101},
URL = { 
        https://doi.org/10.1080/03610927408827101
},
}

@ARTICLE{dbindex,
  author={Davies, David L. and Bouldin, Donald W.},
  journal={IEEE Transactions on Pattern Analysis and Machine Intelligence}, 
  title={{A Cluster Separation Measure}}, 
  year={1979},
  volume={PAMI-1},
  number={2},
  pages={224-227},
  doi={10.1109/TPAMI.1979.4766909}}

@article{silhouette,
title = {Silhouettes: A graphical aid to the interpretation and validation of cluster analysis},
journal = {Journal of Computational and Applied Mathematics},
volume = {20},
pages = {53-65},
year = {1987},
issn = {0377-0427},
doi = {https://doi.org/10.1016/0377-0427(87)90125-7},
url = {https://www.sciencedirect.com/science/article/pii/0377042787901257},
author = {Peter J. Rousseeuw}
}

@article{McInnes2018, 
doi = {10.21105/joss.00861}, 
url = {https://doi.org/10.21105/joss.00861},
year = {2018}, 
publisher = {The Open Journal}, 
volume = {3}, 
number = {29}, 
pages = {861}, 
author = {McInnes, Leland and Healy, John and Saul, Nathaniel and Großberger, Lukas}, 
title = {{UMAP: Uniform Manifold Approximation and Projection}}, 
journal = {Journal of Open Source Software} 
}

@article{hungarian,
author = {Kuhn, Harold W.},
title = {{The Hungarian method for the assignment problem}},
journal = {Naval Research Logistics Quarterly},
volume = {2},
number = {1-2},
pages = {83-97},

url = {https://onlinelibrary.wiley.com/doi/abs/10.1002/nav.3800020109},
year = {1955}
}

@misc{elhage_toy_2022,
  title = {Toy {{models}} of {{superposition}}},
  author = {Elhage, Nelson and Hume, Tristan and Olsson, Catherine and Schiefer, Nicholas and Henighan, Tom and Kravec, Shauna and {Hatfield-Dodds}, Zac and Lasenby, Robert and Drain, Dawn and Chen, Carol and Grosse, Roger and McCandlish, Sam and Kaplan, Jared and Amodei, Dario and Wattenberg, Martin and Olah, Christopher},
  year = {2022},
  month = sep,
  eprint = {2209.10652},
  primaryclass = {cs},
  publisher = {arXiv},
  doi = {10.48550/arXiv.2209.10652},
  urldate = {2024-08-07},
  archiveprefix = {arXiv},
  journal = {arXiv preprint arXiv:2209.10652},
  url={https://arxiv.org/abs/2209.10652}
}

@inproceedings{cunningham_sparse_2023,
title={{Sparse Autoencoders Find Highly Interpretable Features in Language Models}},
author={Robert Huben and Hoagy Cunningham and Logan Riggs Smith and Aidan Ewart and Lee Sharkey},
booktitle={The Twelfth International Conference on Learning Representations},
year={2024},
url={https://openreview.net/forum?id=F76bwRSLeK},
address = {Vienna, Austria},
publisher = {ICLR},
pages = {7827--7845},
}

@inproceedings{ma2024doesvlmclassificationbenefit,
author = {Ma, Pingchuan and Rietdorf, Lennart and Kotovenko, Dmytro and Hu, Vincent Tao and Ommer, Bj\"{o}rn},
title = {{Does VLM classification benefit from LLM description semantics?}},
address = {Philadelphia, Pennsylvania, US},
year = {2025},
isbn = {978-1-57735-897-8},
publisher = {AAAI Press},
url = {https://doi.org/10.1609/aaai.v39i6.32638},
doi = {10.1609/aaai.v39i6.32638},
booktitle = {Proceedings of the Thirty-Ninth AAAI Conference on Artificial Intelligence and Thirty-Seventh Conference on Innovative Applications of Artificial Intelligence and Fifteenth Symposium on Educational Advances in Artificial Intelligence},
articleno = {665},
numpages = {9},
series = {AAAI'25/IAAI'25/EAAI'25}
}

@inproceedings{octomodelteamOctoOpenSourceGeneralist2024,
   AUTHOR    = {Dibya Ghosh AND Homer Rich Walke AND Karl Pertsch AND Kevin Black AND Oier Mees AND Sudeep Dasari AND Joey Hejna AND Tobias Kreiman AND Charles Xu AND Jianlan Luo AND You Liang Tan AND Lawrence Yunliang Chen AND Quan Vuong AND Ted Xiao AND Pannag R Sanketi AND Dorsa Sadigh AND Chelsea Finn AND Sergey Levine}, 
    TITLE     = {{Octo: An Open-Source Generalist Robot Policy}}, 
    BOOKTITLE = {Proceedings of Robotics: Science and Systems}, 
    YEAR      = {2024}, 
    ADDRESS   = {Delft, Netherlands}, 
    MONTH     = {July}, 
    DOI       = {10.15607/RSS.2024.XX.090},
    publisher = {RSS},
    numpages = {13}
}

@inproceedings{blackVisionLanguageActionFlow2024,
title={$\pi_{0.5}$: {a Vision-Language-Action Model with Open-World Generalization}},
author={Kevin Black and Noah Brown and James Darpinian and Karan Dhabalia and Danny Driess and Adnan Esmail and Michael Robert Equi and Chelsea Finn and Niccolo Fusai and Manuel Y. Galliker and Dibya Ghosh and Lachy Groom and Karol Hausman and brian ichter and Szymon Jakubczak and Tim Jones and Liyiming Ke and Devin LeBlanc and Sergey Levine and Adrian Li-Bell and Mohith Mothukuri and Suraj Nair and Karl Pertsch and Allen Z. Ren and Lucy Xiaoyang Shi and Laura Smith and Jost Tobias Springenberg and Kyle Stachowicz and James Tanner and Quan Vuong and Homer Walke and Anna Walling and Haohuan Wang and Lili Yu and Ury Zhilinsky},
booktitle={9th Annual Conference on Robot Learning},
year={2025},
url={https://openreview.net/forum?id=vlhoswksBO}
}

@InProceedings{zhang_large_2024,
    author    = {Zhang, Kaichen and Shen, Yifei and Li, Bo and Liu, Ziwei},
    title     = {{Large Multi-modal Models Can Interpret Features in Large Multi-modal Models}},
    booktitle = {Proceedings of the IEEE/CVF International Conference on Computer Vision (ICCV)},
    month     = {October},
    year      = {2025},
    pages     = {3650-3661},
    publisher = {IEEE},
    address = {Honolulu, Hawaii, US}
}

@misc{liu2024llavanext,
    title={LLaVA-NeXT: Improved reasoning, OCR, and world knowledge},
    url={https://llava-vl.github.io/blog/2024-01-30-llava-next/},
    author={Liu, Haotian and Li, Chunyuan and Li, Yuheng and Li, Bo and Zhang, Yuanhan and Shen, Sheng and Lee, Yong Jae},
    month={January},
    year={2024}
}

@inproceedings{liu2023evalcrafter,
  title = {{{EvalCrafter}}: {{Benchmarking}} and {{Evaluating Large Video Generation Models}}},
  shorttitle = {{{EvalCrafter}}},
  booktitle = {Proceedings of the {{IEEE}}/{{CVF Conference}} on {{Computer Vision}} and {{Pattern Recognition}}},
  author = {Liu, Yaofang and Cun, Xiaodong and Liu, Xuebo and Wang, Xintao and Zhang, Yong and Chen, Haoxin and Liu, Yang and Zeng, Tieyong and Chan, Raymond and Shan, Ying},
  year = 2024,
  pages = {22139--22149},
  urldate = {2026-08-03},
  langid = {english}
}

@article{pach_sparse_2025,
  title = {Sparse {{autoencoders learn monosemantic Features}} in {{vision-language models}}},
  author = {Pach, Mateusz and Karthik, Shyamgopal and Bouniot, Quentin and Belongie, Serge and Akata, Zeynep},
  year = {2025},
  month = apr,
  eprint = {2504.02821},
  primaryclass = {cs},
  publisher = {arXiv},
  doi = {10.48550/arXiv.2504.02821},
  urldate = {2025-04-04},
  archiveprefix = {arXiv},
  langid = {english},
  journal = {{Poster on Neural Information Processing Systems (NeurIPS 2025)}},
  url={https://neurips.cc/virtual/2025/loc/san-diego/poster/119210}
}

@misc{grant2026featurescreatedequalmechanistic,
  title         = {{Not All Features Are Created Equal: A Mechanistic
                   Study of Vision-Language-Action Models}},
  author        = {Bryce Grant and Xijia Zhao and Peng Wang},
  year          = {2026},
  eprint        = {2603.19233},
  archivePrefix = {arXiv},
  primaryClass  = {cs.RO},
  url           = {https://arxiv.org/abs/2603.19233}
}

@inproceedings{belkhaleRTHActionHierarchies2024,
    AUTHOR    = {Suneel Belkhale AND Tianli Ding AND Ted Xiao AND Pierre Sermanet AND Quan Vuong AND Jonathan Tompson AND Yevgen Chebotar AND Debidatta Dwibedi AND Dorsa Sadigh}, 
    TITLE     = {{RT-H: Action Hierarchies using Language}}, 
    BOOKTITLE = {Proceedings of Robotics: Science and Systems}, 
    YEAR      = {2024}, 
    ADDRESS   = {Delft, Netherlands}, 
    MONTH     = {July}, 
    DOI       = {10.15607/RSS.2024.XX.049},
    numpages = {23},
    publisher = {RSS}
}

@inproceedings{shiHiRobotOpenEnded2025,
  title = {Hi {{Robot}}: {{Open-Ended Instruction Following}} with {{Hierarchical Vision-Language-Action Models}}},
  shorttitle = {Hi {{Robot}}},
  booktitle = {Forty-Second {{International Conference}} on {{Machine Learning}}},
  author = {Shi, Lucy Xiaoyang and Ichter, Brian and Equi, Michael Robert and Ke, Liyiming and Pertsch, Karl and Vuong, Quan and Tanner, James and Walling, Anna and Wang, Haohuan and Fusai, Niccolo and {Li-Bell}, Adrian and Driess, Danny and Groom, Lachy and Levine, Sergey and Finn, Chelsea},
  year = 2025,
  month = jun,
  urldate = {2026-02-04},
  langid = {english}
}

@misc{nvidia2025gr00tn1openfoundation,
      title={{GR00T N1: An Open Foundation Model for Generalist Humanoid Robots}}, 
      author={NVIDIA and : and Johan Bjorck and Fernando Castañeda and Nikita Cherniadev and Xingye Da and Runyu Ding and Linxi "Jim" Fan and Yu Fang and Dieter Fox and Fengyuan Hu and Spencer Huang and Joel Jang and Zhenyu Jiang and Jan Kautz and Kaushil Kundalia and Lawrence Lao and Zhiqi Li and Zongyu Lin and Kevin Lin and Guilin Liu and Edith Llontop and Loic Magne and Ajay Mandlekar and Avnish Narayan and Soroush Nasiriany and Scott Reed and You Liang Tan and Guanzhi Wang and Zu Wang and Jing Wang and Qi Wang and Jiannan Xiang and Yuqi Xie and Yinzhen Xu and Zhenjia Xu and Seonghyeon Ye and Zhiding Yu and Ao Zhang and Hao Zhang and Yizhou Zhao and Ruijie Zheng and Yuke Zhu},
      year={2025},
      eprint={2503.14734},
      archivePrefix={arXiv},
      primaryClass={cs.RO},
      url={https://arxiv.org/abs/2503.14734}, 
}

@INPROCEEDINGS{openx2024,
  author={O’Neill, Abby and Rehman, Abdul and Maddukuri, Abhiram and Gupta, Abhishek and Padalkar, Abhishek and Lee, Abraham and Pooley, Acorn and Gupta, Agrim and Mandlekar, Ajay and Jain, Ajinkya and Tung, Albert and Bewley, Alex and Herzog, Alex and Irpan, Alex and Khazatsky, Alexander and Rai, Anant and Gupta, Anchit and Wang, Andrew and Singh, Anikait and Garg, Animesh and Kembhavi, Aniruddha and Xie, Annie and Brohan, Anthony and Raffin, Antonin and Sharma, Archit and Yavary, Arefeh and Jain, Arhan and Balakrishna, Ashwin and Wahid, Ayzaan and Burgess-Limerick, Ben and Kim, Beomjoon and Schölkopf, Bernhard and Wulfe, Blake and Ichter, Brian and Lu, Cewu and Xu, Charles and Le, Charlotte and Finn, Chelsea and Wang, Chen and Xu, Chenfeng and Chi, Cheng and Huang, Chenguang and Chan, Christine and Agia, Christopher and Pan, Chuer and Fu, Chuyuan and Devin, Coline and Xu, Danfei and Morton, Daniel and Driess, Danny and Chen, Daphne and Pathak, Deepak and Shah, Dhruv and Büchler, Dieter and Jayaraman, Dinesh and Kalashnikov, Dmitry and Sadigh, Dorsa and Johns, Edward and Foster, Ethan and Liu, Fangchen and Ceola, Federico and Xia, Fei and Zhao, Feiyu and Stulp, Freek and Zhou, Gaoyue and Sukhatme, Gaurav S. and Salhotra, Gautam and Yan, Ge and Feng, Gilbert and Schiavi, Giulio and Berseth, Glen and Kahn, Gregory and Wang, Guanzhi and Su, Hao and Fang, Hao-Shu and Shi, Haochen and Bao, Henghui and Ben Amor, Heni and Christensen, Henrik I and Furuta, Hiroki and Walke, Homer and Fang, Hongjie and Ha, Huy and Mordatch, Igor and Radosavovic, Ilija and Leal, Isabel and Liang, Jacky and Abou-Chakra, Jad and Kim, Jaehyung and Drake, Jaimyn and Peters, Jan and Schneider, Jan and Hsu, Jasmine and Bohg, Jeannette and Bingham, Jeffrey and Wu, Jeffrey and Gao, Jensen and Hu, Jiaheng and Wu, Jiajun and Wu, Jialin and Sun, Jiankai and Luo, Jianlan and Gu, Jiayuan and Tan, Jie and Oh, Jihoon and Wu, Jimmy and Lu, Jingpei and Yang, Jingyun and Malik, Jitendra and Silvério, João and Hejna, Joey and Booher, Jonathan and Tompson, Jonathan and Yang, Jonathan and Salvador, Jordi and Lim, Joseph J. and Han, Junhyek and Wang, Kaiyuan and Rao, Kanishka and Pertsch, Karl and Hausman, Karol and Go, Keegan and Gopalakrishnan, Keerthana and Goldberg, Ken and Byrne, Kendra and Oslund, Kenneth and Kawaharazuka, Kento and Black, Kevin and Lin, Kevin and Zhang, Kevin and Ehsani, Kiana and Lekkala, Kiran and Ellis, Kirsty and Rana, Krishan and Srinivasan, Krishnan and Fang, Kuan and Singh, Kunal Pratap and Zeng, Kuo-Hao and Hatch, Kyle and Hsu, Kyle and Itti, Laurent and Chen, Lawrence Yunliang and Pinto, Lerrel and Fei-Fei, Li and Tan, Liam and Fan, Linxi Jim and Ott, Lionel and Lee, Lisa and Weihs, Luca and Chen, Magnum and Lepert, Marion and Memmel, Marius and Tomizuka, Masayoshi and Itkina, Masha and Castro, Mateo Guaman and Spero, Max and Du, Maximilian and Ahn, Michael and Yip, Michael C. and Zhang, Mingtong and Ding, Mingyu and Heo, Minho and Srirama, Mohan Kumar and Sharma, Mohit and Kim, Moo Jin and Kanazawa, Naoaki and Hansen, Nicklas and Heess, Nicolas and Joshi, Nikhil J and Suenderhauf, Niko and Liu, Ning and Di Palo, Norman and Shafiullah, Nur Muhammad Mahi and Mees, Oier and Kroemer, Oliver and Bastani, Osbert and Sanketi, Pannag R and Miller, Patrick Tree and Yin, Patrick and Wohlhart, Paul and Xu, Peng and Fagan, Peter David and Mitrano, Peter and Sermanet, Pierre and Abbeel, Pieter and Sundaresan, Priya and Chen, Qiuyu and Vuong, Quan and Rafailov, Rafael and Tian, Ran and Doshi, Ria and Martín-Martín, Roberto and Baijal, Rohan and Scalise, Rosario and Hendrix, Rose and Lin, Roy and Qian, Runjia and Zhang, Ruohan and Mendonca, Russell and Shah, Rutav and Hoque, Ryan and Julian, Ryan and Bustamante, Samuel and Kirmani, Sean and Levine, Sergey and Lin, Shan and Moore, Sherry and Bahl, Shikhar and Dass, Shivin and Sonawani, Shubham and Song, Shuran and Xu, Sichun and Haldar, Siddhant and Karamcheti, Siddharth and Adebola, Simeon and Guist, Simon and Nasiriany, Soroush and Schaal, Stefan and Welker, Stefan and Tian, Stephen and Ramamoorthy, Subramanian and Dasari, Sudeep and Belkhale, Suneel and Park, Sungjae and Nair, Suraj and Mirchandani, Suvir and Osa, Takayuki and Gupta, Tanmay and Harada, Tatsuya and Matsushima, Tatsuya and Xiao, Ted and Kollar, Thomas and Yu, Tianhe and Ding, Tianli and Davchev, Todor and Zhao, Tony Z. and Armstrong, Travis and Darrell, Trevor and Chung, Trinity and Jain, Vidhi and Vanhoucke, Vincent and Zhan, Wei and Zhou, Wenxuan and Burgard, Wolfram and Chen, Xi and Wang, Xiaolong and Zhu, Xinghao and Geng, Xinyang and Liu, Xiyuan and Liangwei, Xu and Li, Xuanlin and Lu, Yao and Ma, Yecheng Jason and Kim, Yejin and Chebotar, Yevgen and Zhou, Yifan and Zhu, Yifeng and Wu, Yilin and Xu, Ying and Wang, Yixuan and Bisk, Yonatan and Cho, Yoonyoung and Lee, Youngwoon and Cui, Yuchen and Cao, Yue and Wu, Yueh-Hua and Tang, Yujin and Zhu, Yuke and Zhang, Yunchu and Jiang, Yunfan and Li, Yunshuang and Li, Yunzhu and Iwasawa, Yusuke and Matsuo, Yutaka and Ma, Zehan and Xu, Zhuo and Cui, Zichen Jeff and Zhang, Zichen and Lin, Zipeng},
  booktitle={2024 IEEE International Conference on Robotics and Automation (ICRA)}, 
  title={{Open X-Embodiment: Robotic Learning Datasets and RT-X Models : Open X-Embodiment Collaboration}}, 
  year={2024},
  volume={},
  number={},
  pages={6892-6903},
  doi={10.1109/ICRA57147.2024.10611477},
  publisher = {IEEE},
  address = {Yokohama, Japan}}

@article{stephensDealingLabelSwitching2000,
  title = {Dealing {{With Label Switching}} in {{Mixture Models}}},
  author = {Stephens, Matthew},
  year = 2000,
  month = nov,
  journal = {Journal of the Royal Statistical Society Series B: Statistical Methodology},
  volume = {62},
  number = {4},
  pages = {795--809},
  issn = {1369-7412, 1467-9868},
  doi = {10.1111/1467-9868.00265},
  urldate = {2026-04-10},
  copyright = {https://academic.oup.com/journals/pages/open\_access/funder\_policies/chorus/standard\_publication\_model},
  langid = {english}
}

@inproceedings{andeol_holistic_2023,
author = {Fel, Thomas and Boutin, Victor and Moayeri, Mazda and Cad\`{e}ne, R\'{e}mi and Bethune, Louis and And\'{e}ol, L\'{e}o and Chalvidal, Mathieu and Serre, Thomas},
title = {A holistic approach to unifying automatic concept extraction and concept importance estimation},
year = {2023},
publisher = {Curran Associates Inc.},
address = {Red Hook, NY, USA},
booktitle = {Proceedings of the 37th International Conference on Neural Information Processing Systems},
articleno = {2391},
numpages = {14},
pages = {54805--54818},
location = {New Orleans, LA, USA},
series = {NIPS '23}
}

@INPROCEEDINGS{peebles2023dit,
author = { Peebles, William and Xie, Saining },
booktitle = { 2023 IEEE/CVF International Conference on Computer Vision (ICCV) },
title = {{ Scalable Diffusion Models with Transformers }},
year = {2023},
volume = {},
ISSN = {},
pages = {4172-4182},
doi = {10.1109/ICCV51070.2023.00387},
url = {https://doi.ieeecomputersociety.org/10.1109/ICCV51070.2023.00387},
publisher = {IEEE Computer Society},
address = {Los Alamitos, CA, USA},
month =Oct}

@inproceedings{hawaslyScalingDiscoveryLatent2024,
  title = {Scaling up {{Discovery}} of {{Latent Concepts}} in {{Deep NLP Models}}},
  booktitle = {Proceedings of the 18th {{Conference}} of the {{European Chapter}} of the {{Association}} for {{Computational Linguistics}} ({{Volume}} 1: {{Long Papers}})},
  author = {Hawasly, Majd and Dalvi, Fahim and Durrani, Nadir},
  editor = {Graham, Yvette and Purver, Matthew},
  year = 2024,
  month = mar,
  pages = {793--806},
  publisher = {Association for Computational Linguistics},
  address = {St. Julian's, Malta},
  doi = {10.18653/v1/2024.eacl-long.48},
  urldate = {2026-04-23}
}

@article{li2025eagle2,
    title={{Eagle 2: Building Post-Training Data Strategies from Scratch for Frontier Vision-Language Models}}, 
    author={Zhiqi Li and Guo Chen and Shilong Liu and Shihao Wang and Vibashan VS and Yishen Ji and Shiyi Lan and Hao Zhang and Yilin Zhao and Subhashree Radhakrishnan and Nadine Chang and Karan Sapra and Amala Sanjay Deshmukh and Tuomas Rintamaki and Matthieu Le and Ilia Karmanov and Lukas Voegtle and Philipp Fischer and De-An Huang and Timo Roman and Tong Lu and Jose M. Alvarez and Bryan Catanzaro and Jan Kautz and Andrew Tao and Guilin Liu and Zhiding Yu},
    journal={arXiv:2501.14818},
    year={2025}
}

@inproceedings{haonMechanisticInterpretabilitySteering2025a,
  title = {Mechanistic {{Interpretability}} for {{Steering Vision-Language-Action Models}}},
  booktitle = {9th {{Annual Conference}} on {{Robot Learning}}},
  author = {H{\"a}on, Bear and Stocking, Kaylene Caswell and Chuang, Ian and Tomlin, Claire},
  year = 2025,
  month = sep,
  urldate = {2026-07-28},
  langid = {english}
}
\newpage
\appendix

\begin{figure}[t]
    \centering
    \begin{subfigure}{0.4\textwidth}
        \centering
        \includegraphics[trim={0 0 0 1.2cm}, clip, width=\linewidth]{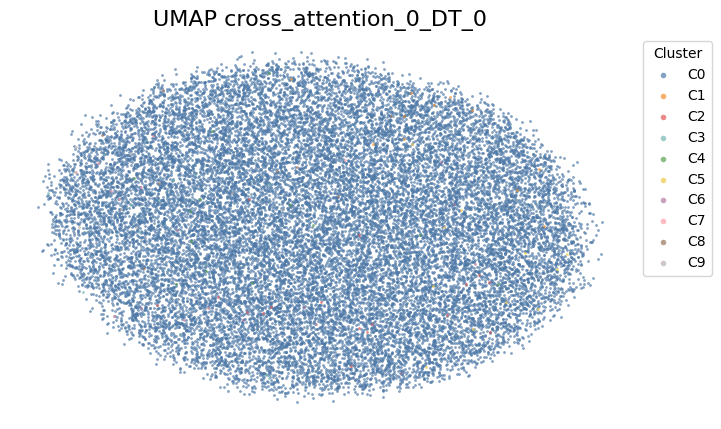}
        \caption{UMAP of PCA compressed embeddings in layer 0 at diffusion timestep 0.}
        \label{fig:pca_umap_ca0_0}
    \end{subfigure}
    \begin{subfigure}{0.4\textwidth}
        \centering
        \includegraphics[trim={0 0 0 1.2cm}, clip, width=\linewidth]{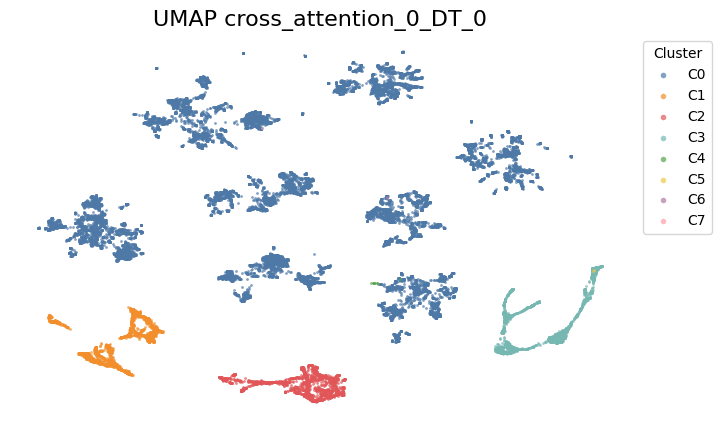}
        \caption{UMAP of weighted and PCA compressed embeddings in layer 0 at diffusion timestep 0.}
        \label{fig:weighted_pca_umap_ca0_0}
    \end{subfigure}
        \begin{subfigure}{0.4\textwidth}
        \centering
        \includegraphics[trim={0 0 0 1.2cm}, clip, width=\linewidth]{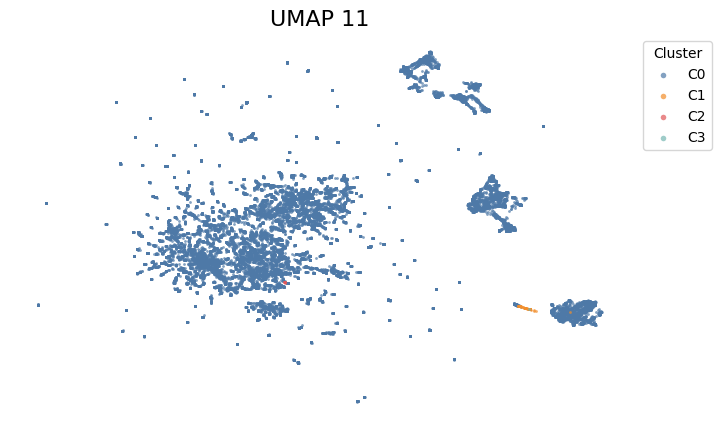}
        \caption{UMAP of embeddings from VLM backbone layer 12 (no diffusion timestep).}
        \label{fig:pca_umap11}
    \end{subfigure}
    \begin{subfigure}{0.4\textwidth}
        \centering
        \includegraphics[trim={0 0 0 1.2cm}, clip, width=\linewidth]{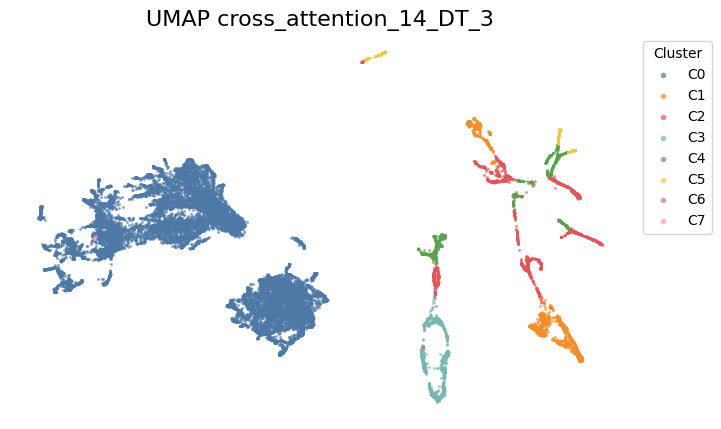}
        \caption{UMAP of PCA compressed embeddings in layer 14 at diffusion timestep 3.}
        \label{fig:pca_umap_ca14_3}
    \end{subfigure}
    \begin{subfigure}{0.4\textwidth}
        \centering
        \includegraphics[trim={0 0 0 1.2cm}, clip, width=\linewidth]{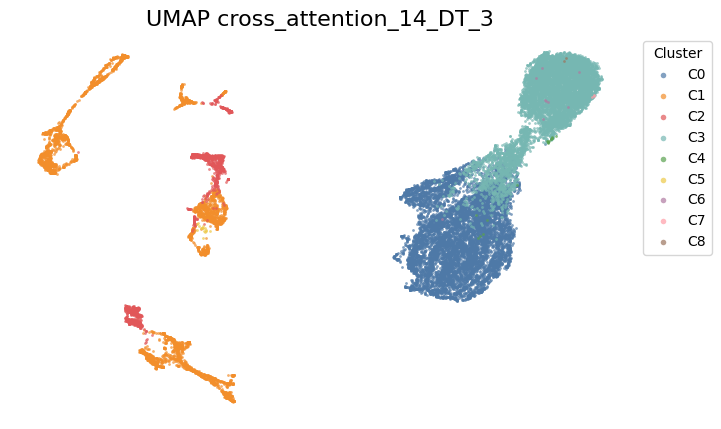}
        \caption{UMAP of weighted and PCA compressed embeddings in layer 14 at diffusion timestep 3.}
        \label{fig:weighted_pca_umap_ca14_3}
    \end{subfigure}
\caption{Comparative UMAP visualisations of the latent space at different layers and diffusion timesteps using PCA-compressed embeddings without and with a weighting mechanism.}
\label{fig:umap}
\end{figure}
\section{Data}
\label{app:data}
We collect embeddings from a part of the Open-X-Embodiment dataset featuring a GR1 robot performing humanoid robot tabletop manipulation.
In total, we have 240k trajectories across 24 tasks at our disposal with visual observations from the ego perspective of the GR1 robot~\footnote{Fourier GR-1 robot: \href{https://www.fftai.com/products-gr1}{https://www.fftai.com/products-gr1}}. 
Due to memory constraints, we load 1000 samples per task, resulting in a final dataset of 24k observation-action pairs. 
These samples are used to collect embeddings via the GR00T model. 
Each task of the original dataset contains 10,000 samples, distributed across episodes of varying lengths. 
To spread our selection evenly across different episodes we include every 10th sample in our selection.

Consistent with standard VLA architectures~\cite{nvidia2025gr00tn1openfoundation,kim2025openvla}, the observation consists of the current frame, the task instruction in natural language, and a robot state vector containing the current joint configuration. 
The ground-truth action is normally used for behaviour cloning during training of VLAs, and not required in our investigation.
The state information contains the joint configuration of the hands, arms, neck, and waist joints.
The tasks span a variety of tabletop manipulation tasks involving the picking up and placing down of different objects.
We include all 24 tasks in our investigation.

\section{Cluster Alignment}
\label{app:alignment}
The clusters are computed independently for each layer and diffusion timestep combination (if applicable).
This results in clusters between layers possibly containing the same or similar samples but having different cluster identifiers assigned.
This phenomenon is known as label switching~\cite{stephensDealingLabelSwitching2000}.
To ensure consistent assignment of cluster identifiers, an additional alignment step is required.
We resolve the label switching occurring between cluster assignments of different layers and diffusion timesteps by performing optimal bipartite matching with the Hungarian algorithm~\cite{hungarian}.
By using the degree of sample overlap between clusters to define the cost matrix, we can consistently trace the evolution of cluster memberships as data flows through the model.

\clearpage
\begin{figure*}[t]
\centering
\setlength{\tabcolsep}{2pt} 
\renewcommand{\arraystretch}{1.0}
\begin{tabular}{cccc}
    \subcaptionbox{Layer 0, Timestep 0}[0.21\textwidth]{%
        \includegraphics[width=0.21\textwidth]{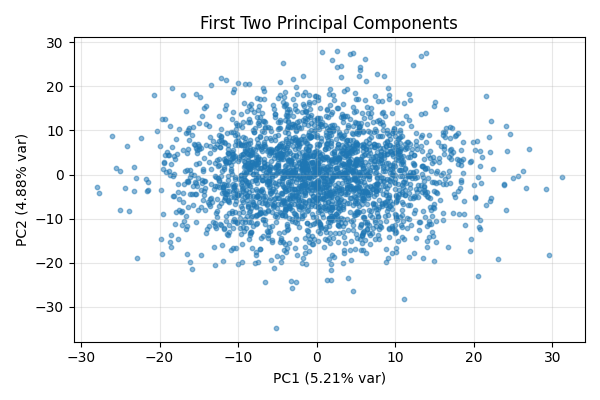}} &
    \subcaptionbox{Layer 0, Timestep 1}[0.21\textwidth]{%
        \includegraphics[width=0.21\textwidth]{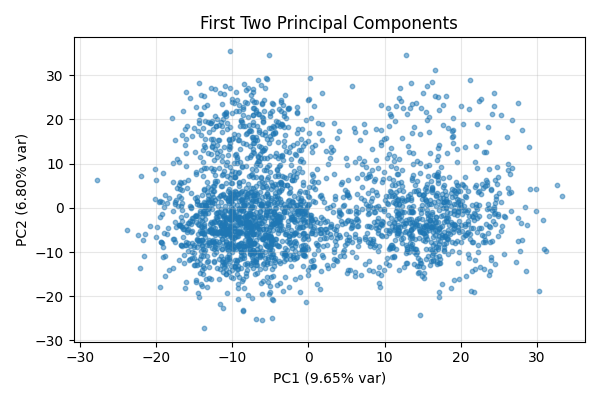}} &
    \subcaptionbox{Layer 0, Timestep 2}[0.21\textwidth]{%
        \includegraphics[width=0.21\textwidth]{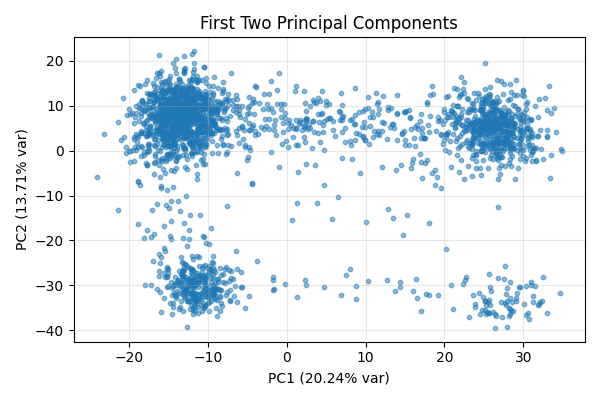}} &
    \subcaptionbox{Layer 0, Timestep 3}[0.21\textwidth]{%
        \includegraphics[width=0.21\textwidth]{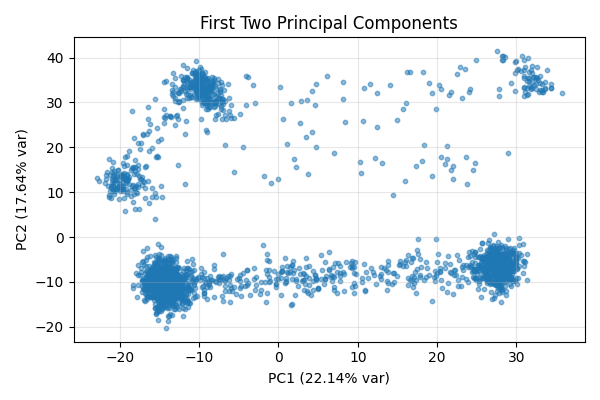}} \\[4pt]
    
    \subcaptionbox{Layer 6, Timestep 0}[0.21\textwidth]{%
        \includegraphics[width=0.21\textwidth]{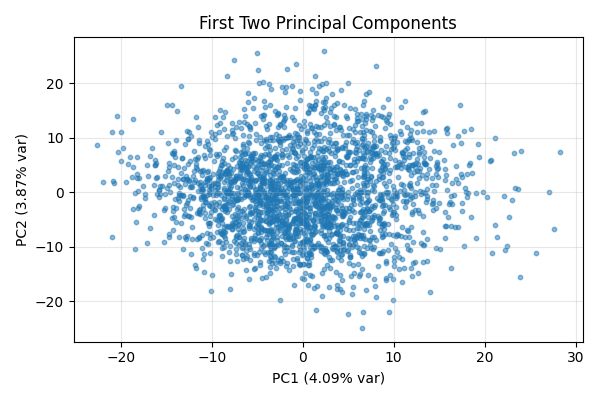}} &
    \subcaptionbox{Layer 6, Timestep 1}[0.21\textwidth]{%
        \includegraphics[width=0.21\textwidth]{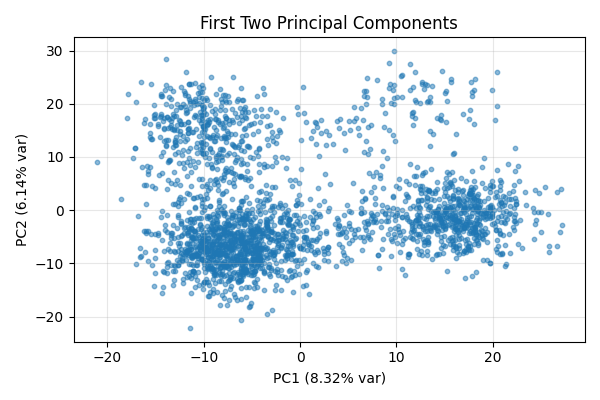}} &
    \subcaptionbox{Layer 6, Timestep 2}[0.21\textwidth]{%
        \includegraphics[width=0.21\textwidth]{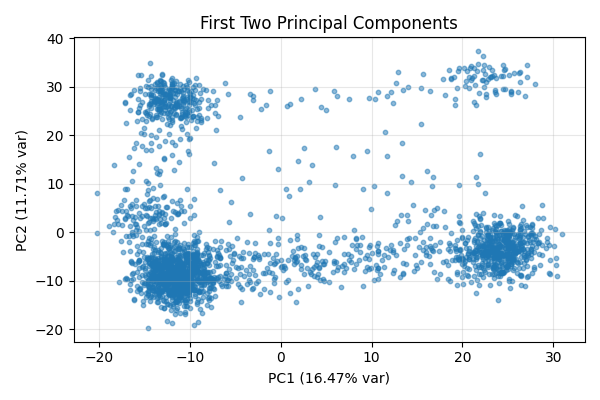}} &
    \subcaptionbox{Layer 6, Timestep 3}[0.21\textwidth]{%
        \includegraphics[width=0.21\textwidth]{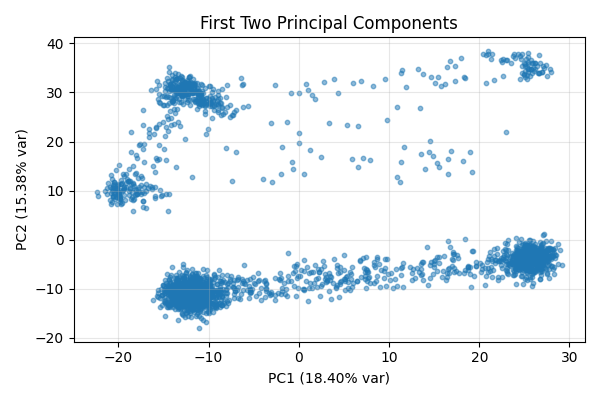}} \\[4pt]
    
    \subcaptionbox{Layer 14, Timestep 0}[0.21\textwidth]{%
        \includegraphics[width=0.21\textwidth]{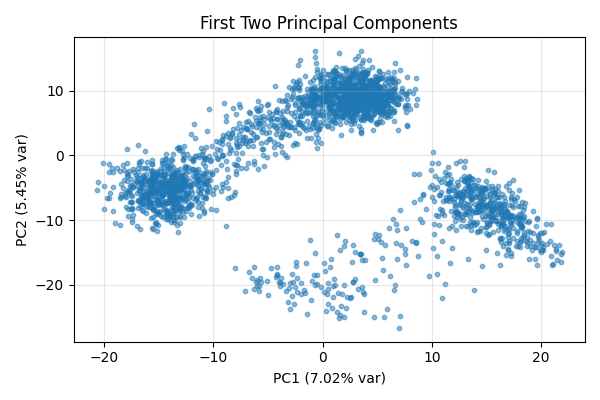}} &
    \subcaptionbox{Layer 14, Timestep 1}[0.21\textwidth]{%
        \includegraphics[width=0.21\textwidth]{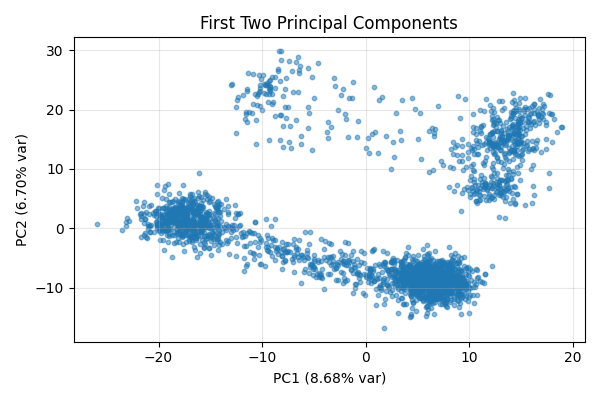}} &
    \subcaptionbox{Layer 14, Timestep 2}[0.21\textwidth]{%
        \includegraphics[width=0.21\textwidth]{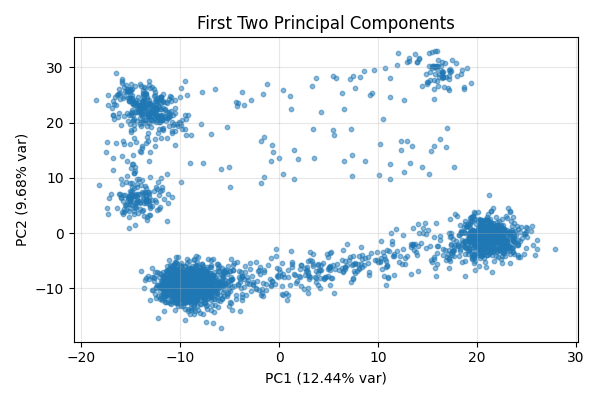}} &
    \subcaptionbox{Layer 14, Timestep 3}[0.21\textwidth]{%
        \includegraphics[width=0.21\textwidth]{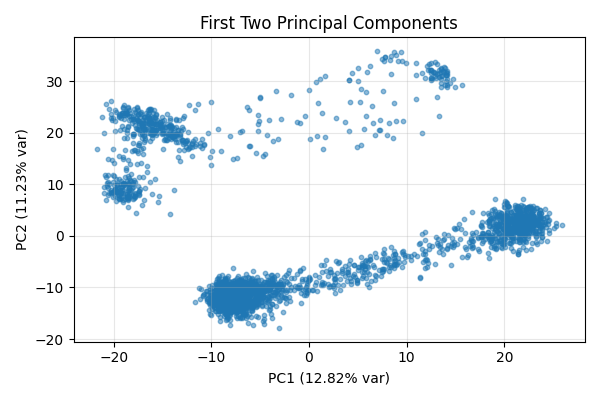}} \\[4pt]
    
    \subcaptionbox{Layer 15, Timestep 0}[0.21\textwidth]{%
        \includegraphics[width=0.21\textwidth]{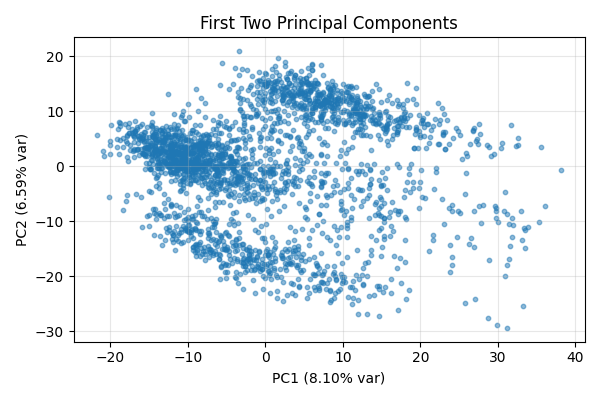}} &
    \subcaptionbox{Layer 15, Timestep 1}[0.21\textwidth]{%
        \includegraphics[width=0.21\textwidth]{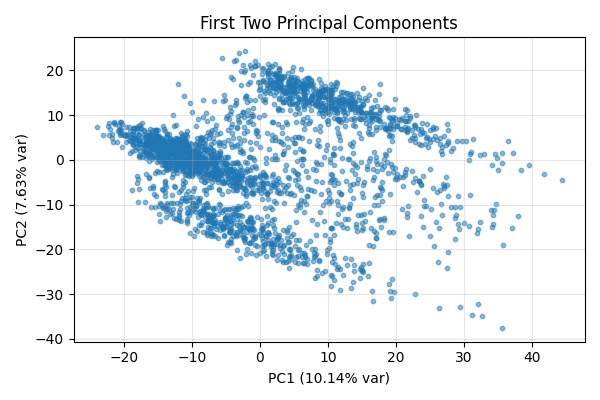}} &
    \subcaptionbox{Layer 15, Timestep 2}[0.21\textwidth]{%
        \includegraphics[width=0.21\textwidth]{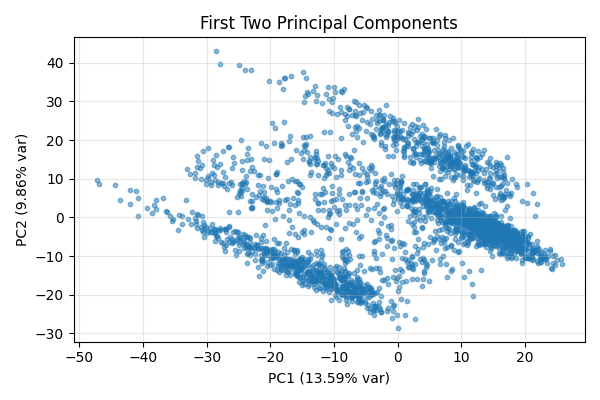}} &
    \subcaptionbox{Layer 15, Timestep 3}[0.21\textwidth]{%
        \includegraphics[width=0.21\textwidth]{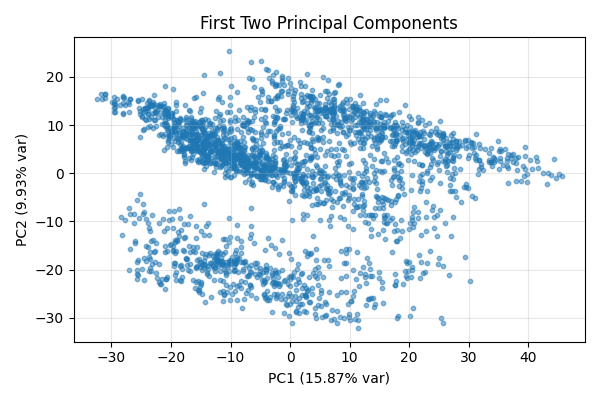}} \\
\end{tabular}
\caption{Visualisation of the first two principal components of the latent space across layers (rows) and timesteps (columns). Each subplot shows the latent space structure for a specific layer–timestep combination, illustrating how latent representations evolve through the network.}
\label{fig:latent_pca_grid}
\end{figure*}
\section{Formulas}
\label{appendix:formulas}
The Overlap Coefficient $O(A, B)$ is computed as follows:
\begin{equation}
O(A, B) = \begin{cases}
    0 & \text{if } \min(|A|, |B|) = 0\\
    \frac{|A \cap B|}{\min(|A|, |B|)} & else
\end{cases}
\label{eq:overlap}
\end{equation}
Whereas the Jaccard Index $J(A, B)$ is computed as displayed in equation~\ref{eq:jaccard}:
\begin{equation}
J(A, B) = \begin{cases}
    0 & \text{if } |A \cup B| = 0 \\
    \frac{|A \cap B|}{|A \cup B|} & \text{else}
\end{cases}
\label{eq:jaccard}
\end{equation}
This gives us a quantifiable way of measuring how much the internal representations of the model change over different layers and diffusion steps.

\section{Cluster Visualisation}
\label{app:umap}
\textbf{The embedding-weighting module uncovers structural patterns during the initial stages of the diffusion process.}
The UMAP~\cite{McInnes2018} visualisations in Figures~\ref{fig:pca_umap_ca0_0} through~\ref{fig:weighted_pca_umap_ca14_3} demonstrate how the weighting mechanism structures the VLA latent space.
This effect is particularly evident in the early layers, where the weighted embeddings form significantly more coherent clusters than their unweighted counterparts.
The first cross-attention layer at the first diffusion timestep takes as input barely processed noisy action token representations, which leads to the unstructured, circular-shaped latent space in Figure~\ref{fig:pca_umap_ca0_0}. 
In contrast, the UMAP visualisation of the embeddings of the last layer of the VLM display a more ordered latent space. 
As our embedding-weighting mechanism directly incorporates the influence of the VLM embeddings, we are able to improve the ordering of the latent space in early noisy embeddings, based on the actual attention of the model.

\section{PCA Analysis}
\label{app:pca}
\begin{figure}[ht]
    \centering
    \includegraphics[width=0.9\linewidth]{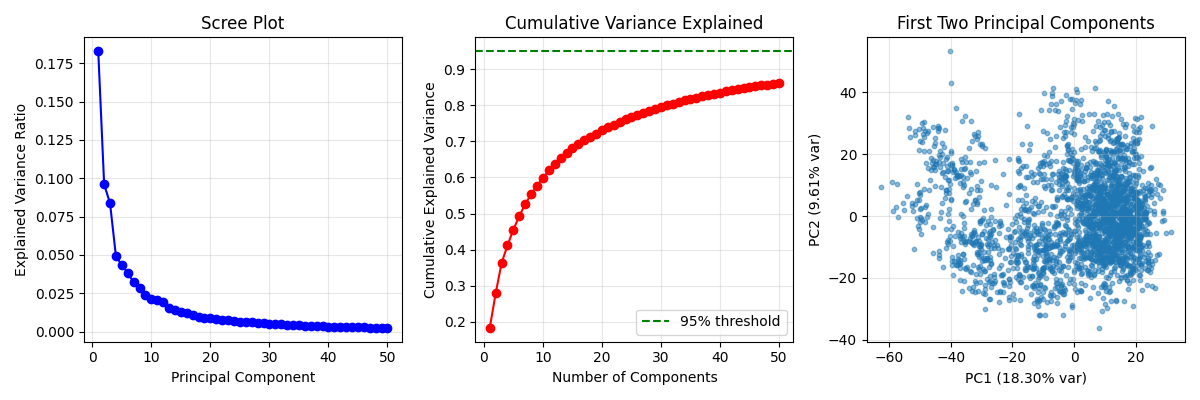}
    \caption{First two principal components of vision-language features produced by VLM backbone.}
    \label{fig:backbone_pca}
\end{figure}
We investigate the principal components of GR00T's internal representations across 10,000 samples from all tasks by visualising the first two principal components and investigating the explained variance of individual components. 
Figure~\ref{fig:backbone_pca} visualises the first two principal components of the vision-language embeddings. 
The Scree plot (left) reveals that the first ten principal components seem to carry the highest amount of explained variance.
However, the plot of the explained cumulative variance (middle) barely exceeds 60\% of explained variance.
This suggests that individual components are responsible only for a small portion of the total information encoded by the single embedding.
The visualisation of the first two components (right) provides some insight into the embedding space, which already provides some recognisable structures.

As can be observed by the visualizations of the first two principal components of the VLA's latent space in Figure~\ref{fig:latent_pca_grid}, the noisy action tokens already show the first signs of clear separability within the first diffusion timestep (see Figure~\ref{fig:latent_pca_grid} (a),(e), and (i)).
Within the same layer, stronger separation occurs over time, which is visible especially across layer 6 (see Figure~\ref{fig:latent_pca_grid} (e) - (h)) where a single noisy cluster splits over time into 5 visually distinct clusters.
Accordingly, the variance of the principal components increases over time.



\section{Hyperparameters}
\label{app:hyperparameters}
Table~\ref{tab:hyperparameters_exhaustive} contains the hyperparameters yielding the best results after sweeping over different distance threshold ranges.
\begin{table}[t]
    \centering
    \caption{Exhaustive list of distance thresholds per layer.}
    \label{tab:hyperparameters_exhaustive}
    \small
    \begin{tabular}{c|c}
    \toprule
        \textbf{Layer} & \textbf{Distance Threshold} \\
        \midrule
        \multicolumn{2}{c}{\textbf{LAVLA (baseline)}} \\\midrule
            VLM Backbone & 9.0383\\ 
            Cross-Attention 0 & 109.1253 \\
            Cross-Attention 2 & 109.9187 \\
            Cross-Attention 4 & 108.6459 \\
            Cross-Attention 6 & 103.7606 \\
            Cross-Attention 8 & 108.2924 \\
            Cross-Attention 10 & 90.0302 \\
            Cross-Attention 12 & 86.9423 \\
            Cross-Attention 14 & 85.0988 \\\midrule 
            
        \multicolumn{2}{c}{\textbf{LAVLA w/ weighted}}  \\\midrule
            VLM Backbone & 9.0383\\ 
            Cross-Attention 0 & 2.6785 \\
            Cross-Attention 2  & 2.7405 \\
            Cross-Attention 4  & 2.9376 \\
            Cross-Attention 6  & 2.8599 \\
            Cross-Attention 8  & 3.6135 \\
            Cross-Attention 10 & 4.1296 \\
            Cross-Attention 12 & 2.8211 \\
            Cross-Attention 14 & 3.2981 \\\midrule 
            
        \multicolumn{2}{c}{\textbf{LAVLA w/ PCA}} \\\midrule
            VLM Backbone & 79.9869\\ 
            Cross-Attention 0 & 60.5738 \\
            Cross-Attention 2 & 60.9242 \\
            Cross-Attention 4 & 59.8374 \\
            Cross-Attention 6 & 61.5705 \\
            Cross-Attention 8 & 56.9619 \\
            Cross-Attention 10 & 59.1317 \\
            Cross-Attention 12 & 54.3040 \\
            Cross-Attention 14 & 47.7121\\\midrule 
            
        \multicolumn{2}{c}{\textbf{LAVLA w/ PCA w/ weighted}} \\\midrule
            VLM Backbone & 79.9869\\ 
            Cross-Attention 0 & 61.3591 \\
            Cross-Attention 2 & 59.0093 \\
            Cross-Attention 4 & 56.0642 \\
            Cross-Attention 6 & 66.1785 \\
            Cross-Attention 8 & 59.5633 \\
            Cross-Attention 10 & 62.7542 \\
            Cross-Attention 12 & 65.4799 \\
            Cross-Attention 14 & 55.7679 \\\bottomrule 
    \end{tabular}
\end{table}

    

\section{Violin plots of top clusters}
\begin{figure}[t]
    \centering
    \begin{subfigure}{0.45\textwidth}
        \centering
        \includegraphics[width=\linewidth]{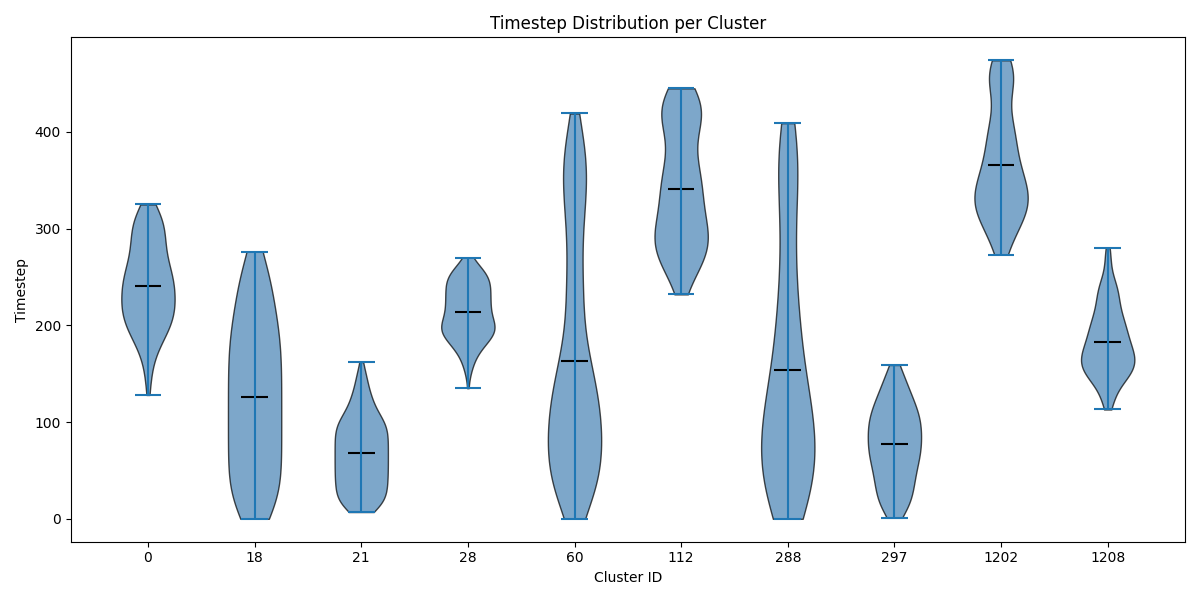}
        \caption{Timesteps}
        \label{fig:bb_timesteps_violin}
    \end{subfigure}
    \begin{subfigure}{0.45\textwidth}
        \centering
        \includegraphics[width=\linewidth]{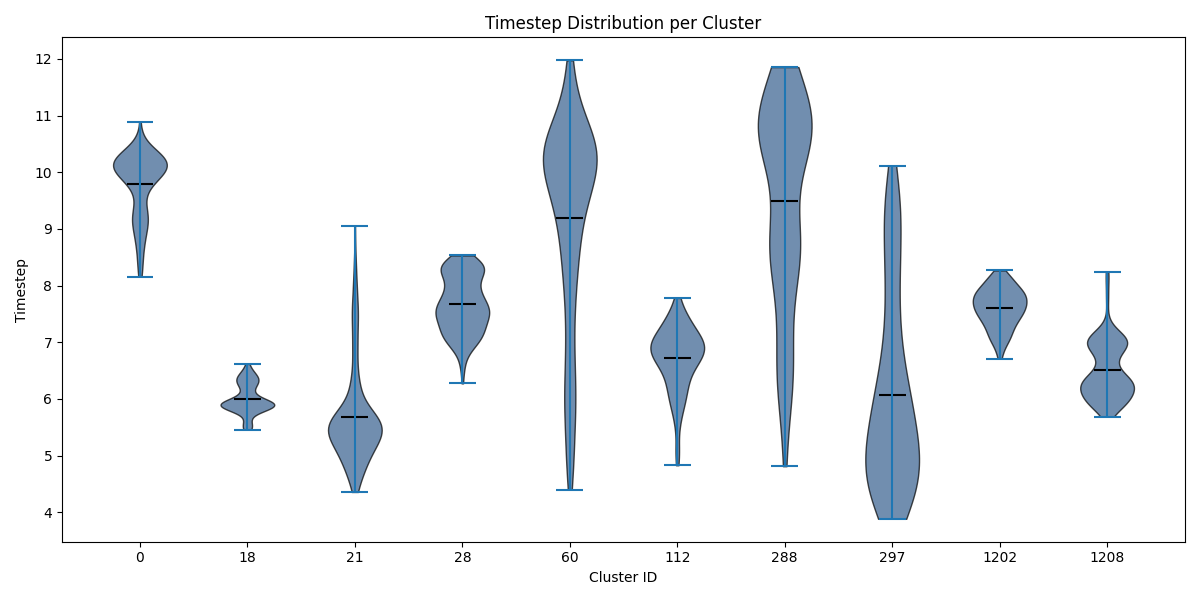}
        \caption{Left arm displacement}
        \label{fig:bb_left_arm_violin}
    \end{subfigure}
    \begin{subfigure}{0.45\textwidth}
        \centering
        \includegraphics[width=\linewidth]{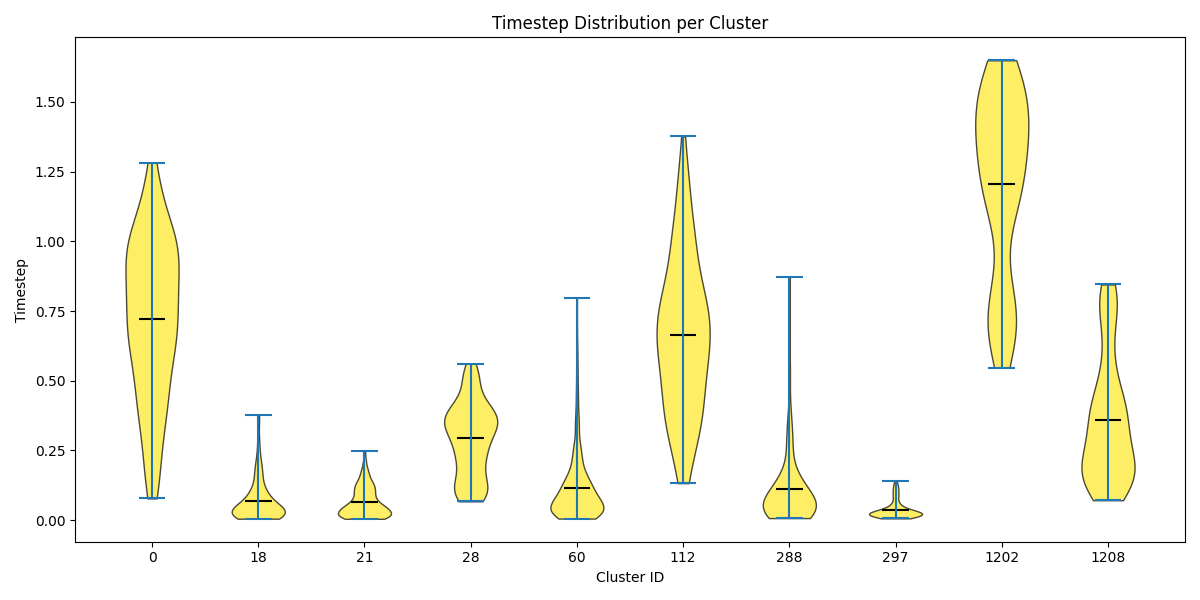}
        \caption{Waist displacement}
        \label{fig:bb_waist_violin}
    \end{subfigure}
    \begin{subfigure}{0.45\textwidth}
        \centering
        \includegraphics[width=\linewidth]{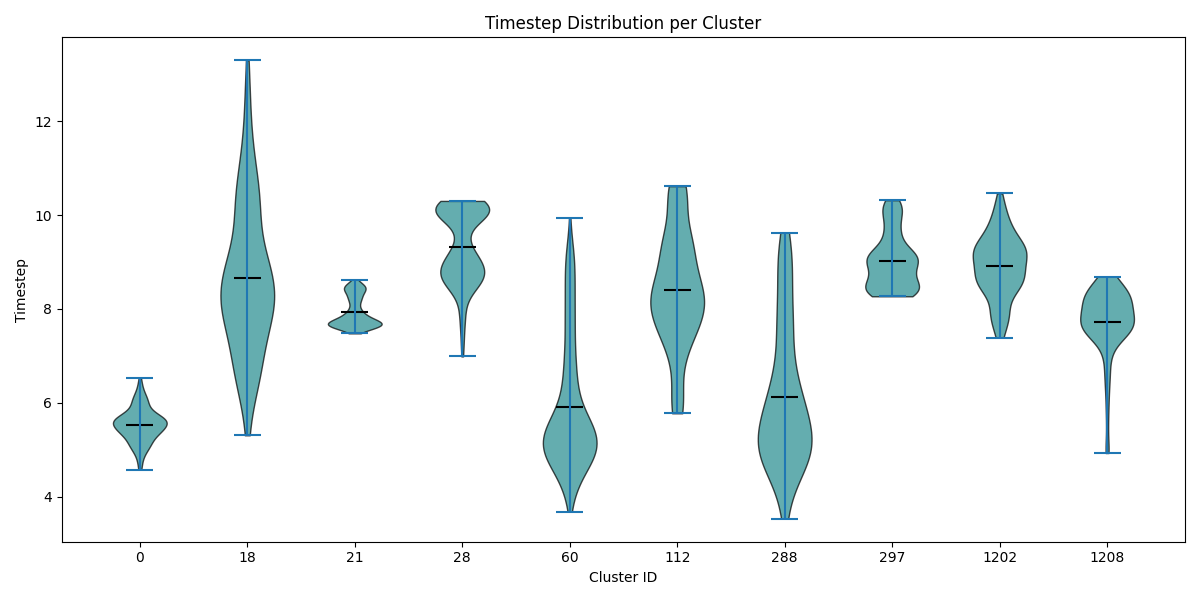}
        \caption{Right arm displacement}
        \label{fig:bb_right_arm_violin}
    \end{subfigure}
    \caption{Violin plots of feature distributions of top clusters within VL backbone}
    \label{fig:violin_bb}
\end{figure}

\begin{figure}[t]
    \centering
    \begin{subfigure}{0.45\textwidth}
        \centering
        \includegraphics[width=\linewidth]{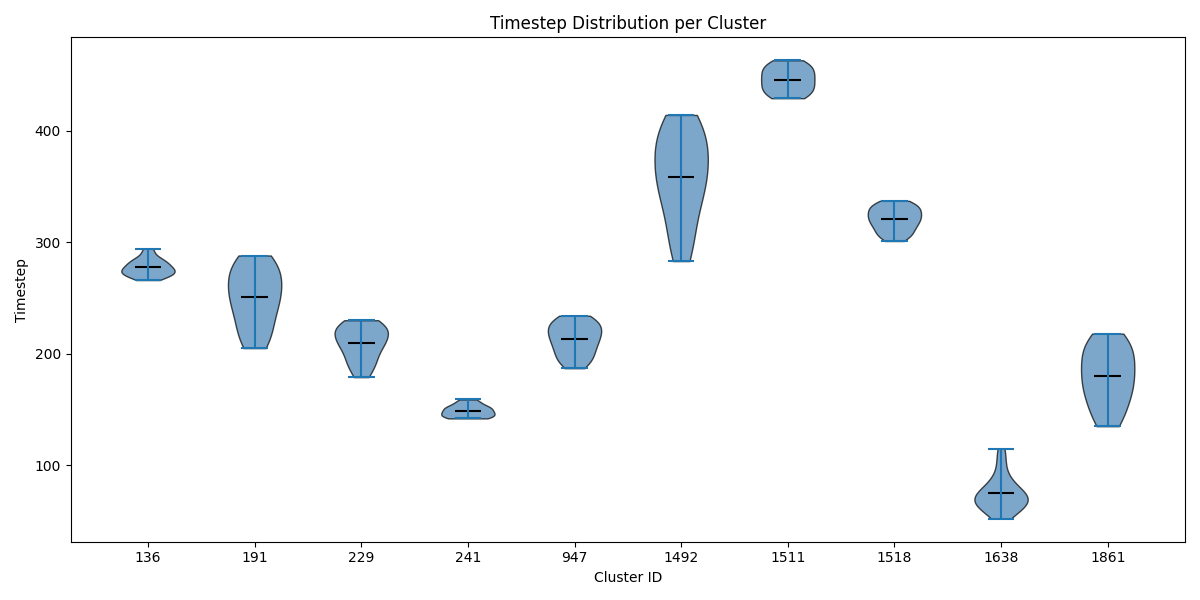}
        \caption{Timesteps}
        \label{fig:ca14_timesteps_violin}
    \end{subfigure}
    \begin{subfigure}{0.45\textwidth}
        \centering
        \includegraphics[width=\linewidth]{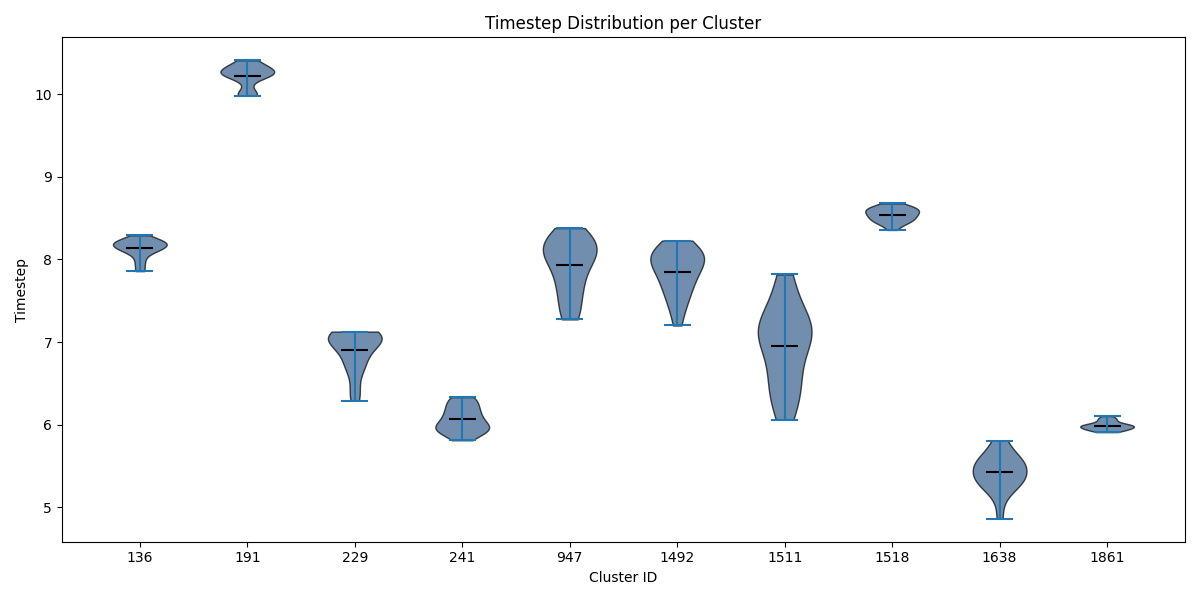}
        \caption{Left arm displacement}
        \label{fig:C14dt3W_left_arm_violin}
    \end{subfigure}
    \begin{subfigure}{0.45\textwidth}
        \centering
        \includegraphics[width=\linewidth]{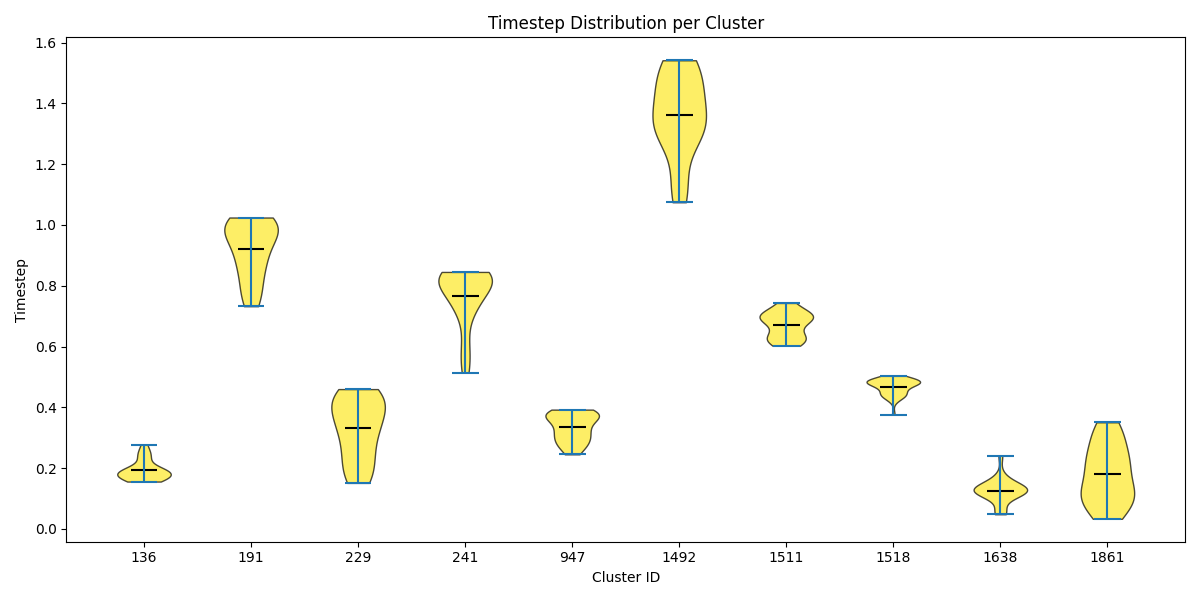}
        \caption{Waist displacement}
        \label{fig:C14dt3W_waist_violin}
    \end{subfigure}
    \begin{subfigure}{0.45\textwidth}
        \centering
        \includegraphics[width=\linewidth]{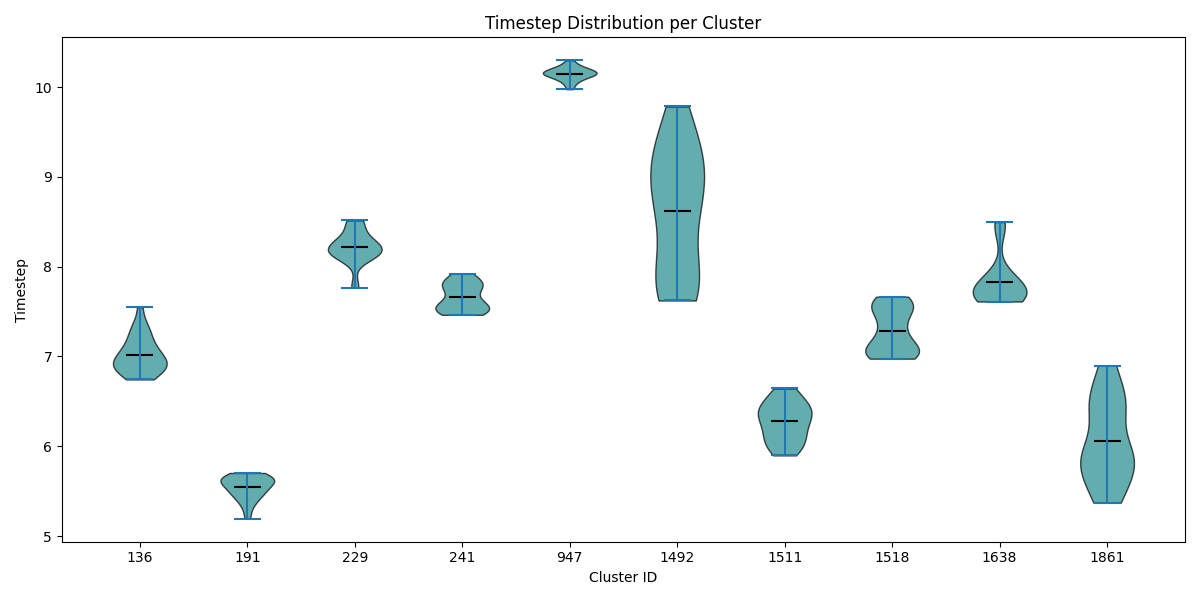}
        \caption{Right arm displacement}
        \label{fig:C14dt3W_right_arm_violin}
    \end{subfigure}
    \caption{Violin plots of feature distributions of top clusters of cross-attention layer 14 at final diffusion timestep (t=3).}
    \label{fig:violin_ca14}
\end{figure}

The Wasserstein distances in Figures~\ref{fig:wasser_bb} and~\ref{fig:wasser_ca14}, provide a quantitative comparison of the different distributions of selected features in the top clusters. Figures~\ref{fig:violin_bb} and~\ref{fig:violin_ca14} provide an additional visualisation of these clusters, highlighting the model's latent space organisation at different layers in the model.
    
\section{Concept generation examples}
\label{app:concept_generation}
\begin{figure*}[!ht]
    \centering
    \begin{subfigure}{\linewidth}
        \centering
        \includegraphics[width=0.8\linewidth]{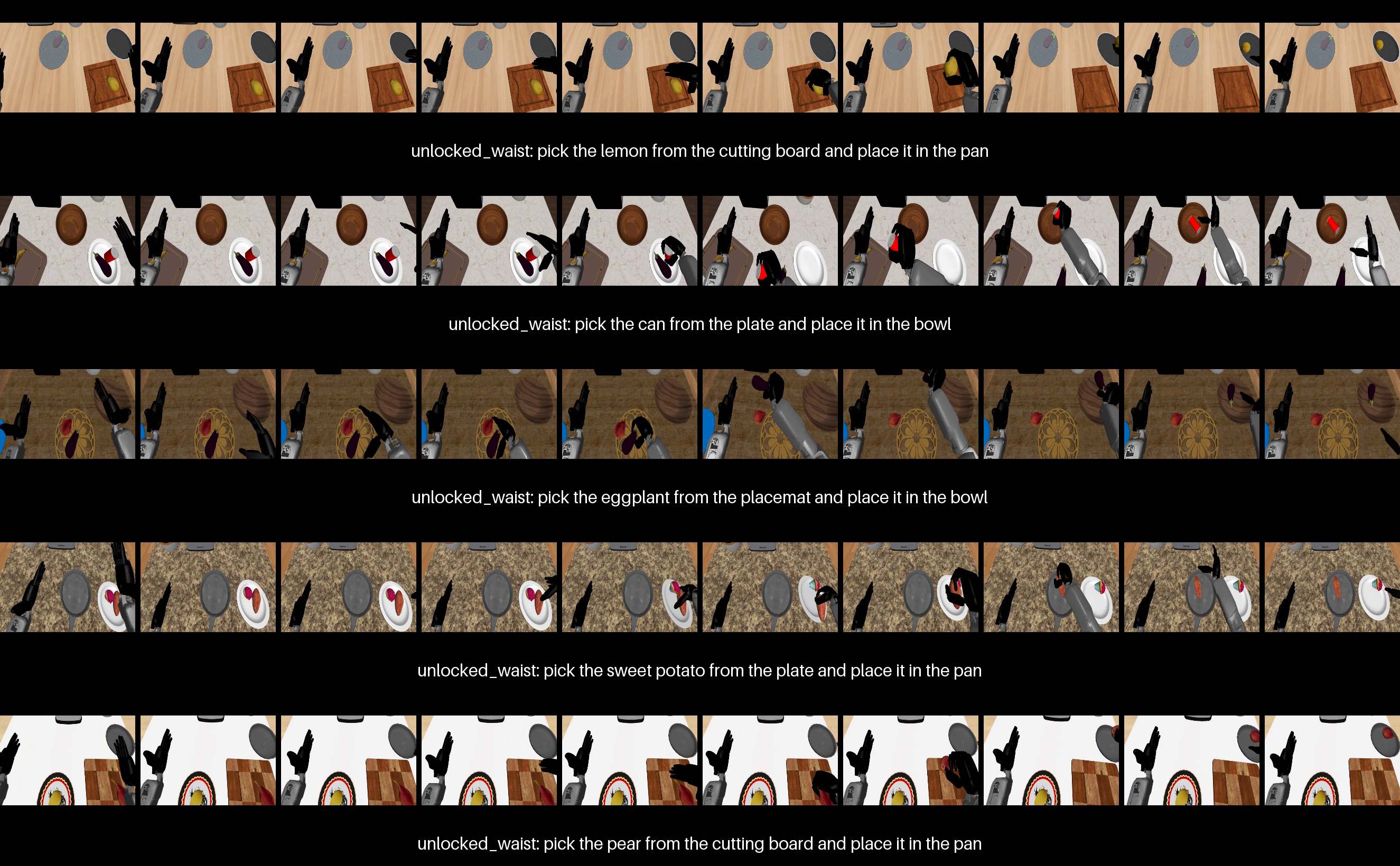}
        \caption{Final concept: The robot's hand picks up an object and places it into a container.}
        \label{fig:concept_generation_1}
    \end{subfigure}

    \vspace{1em}

    \begin{subfigure}{\linewidth}
        \centering
        \includegraphics[width=0.8\linewidth]{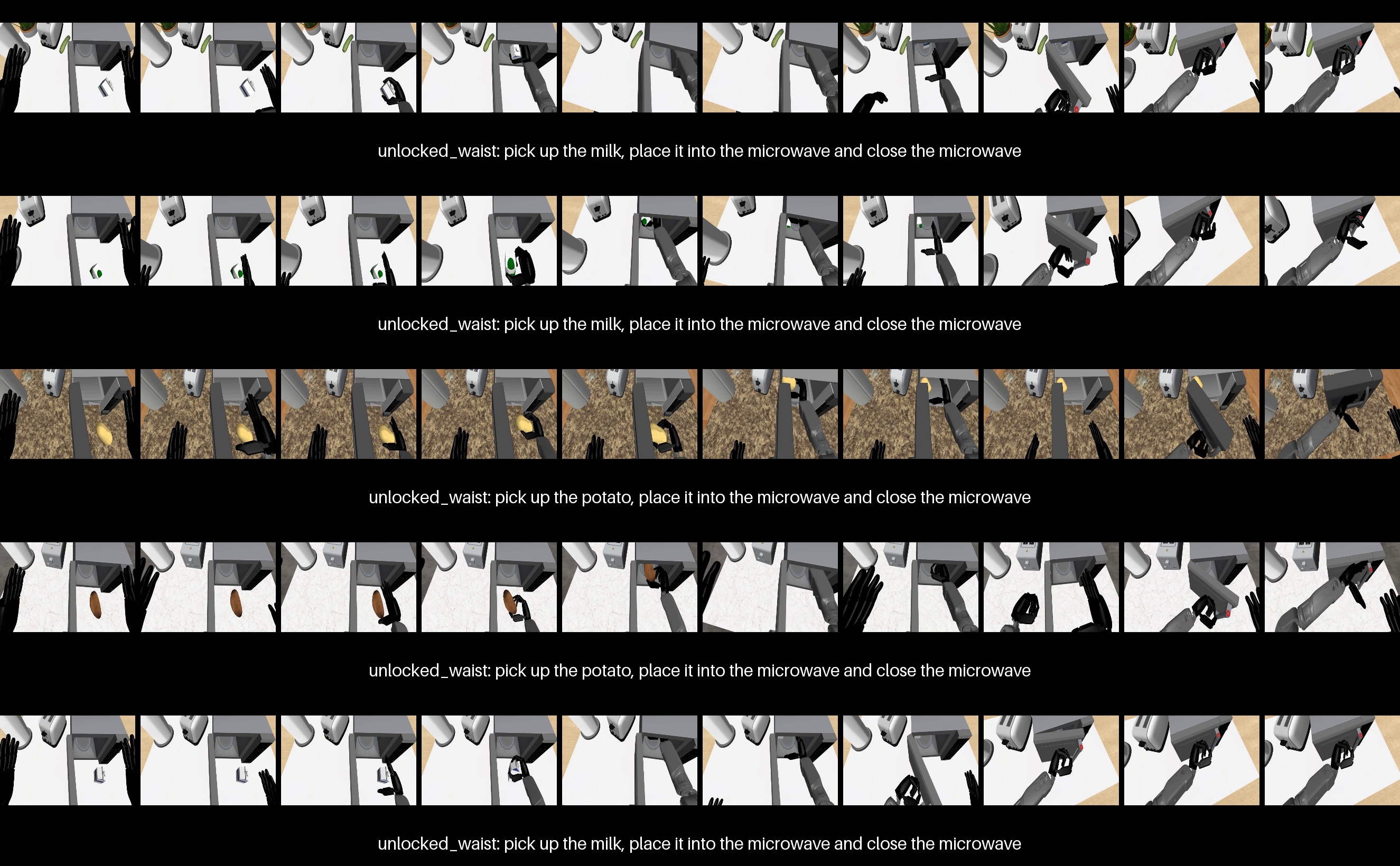}
        \caption{Final concept: The robot picks up an object from a surface and places it into a container.}
        \label{fig:concept_generation_2}
    \end{subfigure}
    \caption{Examples of concept extraction for cross-attention layer 0 using both PCA and weighted clustering.}
    \label{fig:concept_generation}
\end{figure*}

 Figures~\ref{fig:concept_generation_1} and~\ref{fig:concept_generation_2} show two example cluster sample sets used in the concept-generation process. Each concept is derived from 5 videos, each decoded into 10 sequential frames and passed to the VLM (LLaVA-72B) alongside the corresponding task command, shown as the white caption below each image row.

\end{document}